**Title Page**

**Original Article**

# Multi-Large Language Model Orchestrated Severity Assessment of Clinical Records (MOSAIC)

Manuela Del Castillo Suero[1], Arnault-Quentin Vermillet[1], Nicole Sonne Heckmann[1], Darmendra Ramcharran[2,3], Maurizio Sessa[1]

[1]Department of Drug Design and Pharmacology, University of Copenhagen, Copenhagen, Denmark

[2]GSK, Providence, RI, USA

[3]School of Pharmacy, University of Rhode Island, Kingston, RI, USA

**Corresponding author:**
Maurizio Sessa,
Department of Drug Design and Pharmacology,
University of Copenhagen,
Jagtvej 160, 2100 Copenhagen, Denmark
Email: maurizio.sessa@sund.ku.dk

***Abstract***

***Background:*** Disease severity is a multidimensional construct difficult to capture with rule-based approaches in Electronic Healthcare Records (EHR). Agentic large language model (LLM) systems offer a potential approach for synthesising clinical evidence and reasoning over EHRs, but their performance for this task has been yet not evaluated.

***Methods*:** MOSAIC (Multi-agent Orchestrated Severity Assessment In Clinical Records) is a two-phase agentic LLM framework for severity phenotyping, using type 2 diabetes (T2D) as a proof-of-concept. MOSAIC was evaluated on a synthetic EHR cohort (SyntheticMass; open-weight cohort N = 4,886; closed-weight benchmark N = 200) against three established algorithmic ground truths (DCSI GT, DiSSCo, Cooper GT) and by assessing the alignment of its severity classifications with hard clinical outcomes, namely all-cause mortality and incident complications. Open-weight (locally deployable) and proprietary LLM pipelines were also compared.

***Results***: The generated framework extended beyond the domain coverage of the comparators, incorporating biomarker-based glycaemic staging, beta-cell function markers, and social determinants of health not represented in the expert-defined algorithms. Open-weight MOSAIC performed comparably to the proprietary pipeline (closed- vs open-weight $\kappa w = 0.773$) and achieved moderate agreement with Cooper GT ($\kappa w = 0.597$) and DCSI GT ($\kappa w = 0.534$) and fair agreement with DiSSCo GT ($\kappa w = 0.320$). Agent-based (Type 1) severity tiers showed significant separation of all-cause mortality (log-rank $p < 0.001$; crude hazard ratios 1.6–2.4 for non-Baseline tiers), with survival declining across tiers although not strictly monotonically at the upper tiers, and a significant inverse gradient for incident complications (log-rank $p < 0.001$) consistent with depletion of susceptibles in higher-severity patients. Agentic classification also differed substantially from deterministic execution of the same rubric (MOSAIC Frozen; $\kappa w = 0.428$), indicating a contribution of multi-agent reasoning beyond fixed rule execution.

***Conclusion*:** MOSAIC demonstrates that agentic LLM systems can generate and apply clinically meaningful severity phenotypes from structured EHR data when applied to T2D. Broadening the application of MOSAIC to assess severity of other disease phenotypes beyond T2D is an area that warrants further research and evaluation.

## 1. Introduction

The growing availability of longitudinal electronic health records (EHRs) as secondary data sources has expanded opportunities for large-scale pharmacoepidemiology (PE) research, but realising this potential depends on the ability to transform raw clinical information into structured, computable patient representations, a process known as phenotyping (1, 2). Substantial progress has been made in EHR phenotyping using rule-based algorithms, machine learning, and natural language processing (3, 4); however, scalability remains constrained by static representations or extensive manual input, a limitation that becomes particularly problematic when phenotype definitions evolve over time (4, 5). Disease severity is among the most challenging applications of phenotyping: rather than being defined by a single variable, it emerges from the interaction of physiological measurements, complication burden, treatment intensity, and longitudinal disease progression distributed across the EHRs (6). In PE, poorly captured severity phenotypes can introduce bias into treatment comparisons and limit the validity of analyses, whereas robust phenotyping could support better confounding adjustment, risk stratification, and more personalised assessments of treatment response (7-9). Recent advances in large language models (LLMs) offer new possibilities for reasoning over complex, multi-dimensional patient data like EHRs (4, 5, 10), yet their healthcare applications have largely focused on information extraction and classification, with limited exploration of structured phenotyping tasks that require multi-step or longitudinal reasoning, such as severity phenotyping (4). Agentic AI systems, which orchestrate multiple model instances to analyse data and reach conclusions through structured aggregation, offer an even more promising framework for addressing this gap than traditional LLMs (4), however, no studies have yet investigated their performance in severity phenotyping. This study, therefore, aims to develop and evaluate MOSAIC (Multi-agent Orchestrated Severity Assessment In Clinical Records), an agentic multi-LLM framework for severity phenotyping from synthetic longitudinal EHR data, using type 2 diabetes (T2D) as a case study.

## 2. Methods

### *2.1 Data Source*

This study used a high-fidelity synthetic dataset generated by Synthea (11), an open-source patient simulation engine modelling complete patient lifespans via state-transition architectures calibrated to real-world epidemiological sources, including the Global Burden of Disease and NHANES (National Health And Nutrition Examination Survey). The dataset, extracted from SyntheticMass (Version 2, May 2017), comprises approximately one million synthetic patient records in Comma-Separated Values (CSV) format, with clinical terms standardised using Systematized Nomenclature of Medicine – Clinical Terms (SNOMED-CT), RxNorm, and Logical Observation Identifiers Names and Codes (LOINC) ontologies (11), all clinical codes used in the study are listed in Supplementary Table 1. The synthetic data follows a star-schema relational structure in which all clinical tables (i.e., conditions, medications, observations, procedures, immunisations, allergies, and care plans) are linked to a central patient registry via a shared key. Data were available from 23/10/1927 to 27/05/2017.

### *2.2 Study Population and Design*

A new-user cohortdesign was used (12) (Figure 1). To be included in the study population, individuals had to have received a diagnosis of T2D, redeemed their first glucose-lowering treatment following this diagnosis, and survived for at least 5 years from this first treatment (operational definitions provided in Supplementary Table 1). For each patient, the index diagnosis date (t*) was defined as the earliest recorded T2D diagnosis, and the index treatment date (t**) as the first dispensing of a qualifying glucose-lowering agent following t*. A primary index date $T_5$ was defined as t** + 5 years, capturing a clinically distinct disease phase associated with greater complication accumulation and active treatment intensification (13-16). Patients were eligible if both the covariate assessment window [t**, $T_5$] and the five-year follow-up window [$T_5$, $T_5$ + 5 years] fell entirely within the data coverage period (October 1927–May 2017). Severity was classified at the primary index date $T_5$ from EHR data within the covariate assessment window; outcomes were observed during the five-year follow-up, with censoring at death (i.e., when the outcome was not death), last recorded encounter, or end of follow-up, whichever occurred first. The five-year covariate window was selected to capture sufficient disease trajectory and to mirror the assessment horizon of Cooper et al. (2025) (6); the five-year follow-up is consistent with standards for capturing T2D complications and mortality (17).

### *2.3 MOSAIC framework development*

MOSAIC is a two-phase agentic LLM framework for severity phenotyping from longitudinal EHR data. In this study, MOSAIC was applied to T2D.

#### *2.3.1 Phase 1*

In Phase 1, a three-agent CrewAI pipeline synthesised clinical literature to produce a reusable operational T2D severity classification (18). Two search agents, powered by DeepSeek-V3 and GPT-4o were configured as parallel Senior Clinical Researchers and each performed web-based literature retrieval via the Tavily API (*search_depth* = "advanced", *max_results* = 3) (19); this cap was informed by retrieval-augmented generation (RAG) evidence, which showed that retrieval sets of two to four documents optimise downstream generation quality (20-23). A consolidator agent (i.e., Claude Sonnet 4.6) then resolved inter-source discrepancies sequentially and produced a unified four-tier framework, $MOSAIC_F$. $MOSAIC_F$ encoded all severity-classification instructions in a single plain-text (.txt) file, applied uniformly across all patient classifications. Prompting details are provided in Supplementary Table 2.

*2.3.2 Phase 2*

Phase 2 was implemented in two configurations, Type 1 and Type 2.

*Type 1:* In type 1 configuration, for each eligible patient, a plain-text (.txt) file was assembled combining all available EHR data within the covariate window which included demographics, diagnoses, observations, medications, encounters, procedures, and care plans. A three-agent CrewAI pipeline then classified each patient's severity at the primary index date $T_5$ using the operational rubric established in Phase 1 ($MOSAIC_F$). Two of the three agents were assessor agents ($A_1$, $A_2$); each received the patient's text-file record alongside $MOSAIC_F$ and executed asynchronously in parallel, producing independent severity assessments. A consolidator agent ($A_3$) received both assessments, compiled into a single .txt file, as shared context and produced the final tier assignment.

Two architectural configurations were evaluated: a closed-weight (CW; proprietary) pipeline ($A_1$: DeepSeek-V3; $A_2$: GPT-4o; $A_3$: Claude Sonnet 4.6) and an open-weight (OW; open source) pipeline ($A_1$: Gemma 2 27B; $A_2$: Qwen 2.5 14B; $A_3$: Llama 3.1 70B), with OW models served locally via Ollama using 4-bit quantisation. All models were set to temperature 0.0. Full model identifiers, inference settings, and prompt details are provided in Supplementary Tables 3-6. Models were selected based on Best LLMs and Open LLM Leaderboards (24). Although 400B+ models represent the current state of the art, the purposely used OW setup used mid-sized models (14B–70B) compatible with low-computational resources, so that the full ensemble could fit simultaneously, under 4-bit quantisation, on a single GPU. Qwen 2.5 was chosen for structured-data parsing, Gemma 2 for clinical reasoning, and Llama 3.1 (70B) for robust consolidation logic.

*Type 2:* In type 2 configuration, also termed $MOSAIC_{Frozen}$, the research team converted $MOSAIC_F$ into a deterministic rule-based algorithm and applied it directly to the EHR data from the covariate assessment window, allowing severity to be classified without the multi-agent reasoning steps used in Type 1. Type 2 served to isolate the contribution of multi-agent reasoning relative to fixed rule execution in severity classification once $MOSAIC_F$ is retrieved.

### *2.4 Gold standard*

The Type 1 and Type 2 severity-classification configurations (Section 2.3.2) were evaluated for the study outcomes (Section 2.5), against three established algorithmic ground-truth (GT) definitions that served as gold standards. The first GT was the Diabetes Complications Severity Index (DCSI) also known as Young framework (25). The second GT was the Diabetes Severity Score (DiSSCo) (26). The third GT was the framework of Cooper et al. (2025) (6).
For this study, the original diagnostic codes of the three GTs were mapped to SNOMED-CT, LOINC, and RxNorm equivalents, and continuous scores were converted to four ordinal tiers to enable direct comparison with MOSAIC Type 1 and Type 2 severity-classification.

### *2.5 Outcomes*

Four primary outcomes were evaluated. The first was the successful generation of an operational four-tier T2D severity framework. The second was classification agreement between the MOSAIC Type 1 and Type 2 severity classifications and each GT framework. The third was the predictive validity of the MOSAIC severity tiers for downstream outcomes: all-cause mortality, incident T2D complications. The fourth primary outcome was a descriptive presentation of representative examples of disagreement between MOSAIC and GTs.

### *2.6 Statistical Analysis*

#### *2.6.1 Descriptive statistics*

Continuous variables were summarised as mean ± standard deviation (SD) or median [interquartile range, IQR] according to distribution, and categorical variables as counts (percentages). Unless otherwise stated, uncertainty was reported with 95% confidence intervals (CIs) and hypothesis tests were two-sided at $\alpha = 0.05$. Cohort sizes were determined pragmatically rather than by formal power calculation: the CW cohort was restricted to 200 randomly selected patients from the full eligible population owing to API cost and runtime constraints, whereas the OW cohort was expanded to 4,886 randomly selected patients from the full eligible population for the primary scalable analysis owing to computational costs and runtime constraints. Representativeness of the analytic cohorts was assessed by comparison with the full eligible T2D population on sex, race, and age, and between-tier differences in baseline characteristics were quantified using the standardised mean difference (SMD), without formal significance testing of descriptive comparisons. Data completeness and EHR fitness-for-purpose were evaluated descriptively as the proportion of missing values for each dataset field (absolute counts and percentages) and the proportion of patients with at least one qualifying record for every clinical concept required by the frameworks; missingness was interpreted with respect to its clinical mechanism rather than imputed. Person-time, event counts, and crude incidence rates per 100 person-years were tabulated overall and by severity tier, with exact Poisson (chi-squared) 95% CIs.

*2.6.2 Second primary outcome - Classification agreement*

Classification agreement between each MOSAIC configuration (Type 1 CW and OW pipelines; Type 2, $\text{MOSAIC}_{\text{Frozen}}$) and each comparator framework (i.e., GTs) was evaluated at the population and patient levels. At the population level, the distribution of patients across the four severity tiers was modelled with a Bayesian Dirichlet–multinomial model using a symmetric Dirichlet prior (concentration $\alpha = 1$) and a multinomial likelihood, yielding posterior tier probabilities ($\pi$) summarised by their posterior mean and 95% highest-density interval (HDI). The posterior was sampled with the No-U-Turn sampler (four chains, 2,000 draws each after 1,000 warm-up iterations, target acceptance 0.9); convergence was confirmed by $\hat{R} < 1.01$ and effective sample size > 400 for all parameters. Systematic differences in how each framework (i.e., GTs) populated the tiers were assessed via the posterior distribution of the pairwise tier-probability difference ($\Delta_k$ = comparator − OW MOSAIC), a difference being considered credible when its 95% HDI excluded zero.

At the patient level, concordance was quantified using quadratic-weighted Cohen's κ (κw), which penalises disagreements in proportion to the squared distance between tiers, with 95% CIs obtained by bootstrap resampling (2,000 replicates). κw was interpreted against standard benchmarks (≤ 0.20 slight, 0.21–0.40 fair, 0.41–0.60 moderate, 0.61–0.80 substantial, > 0.80 almost perfect) and against pre-specified minimum-acceptable (κw ≥ 0.61) and target (κw ≥ 0.80) thresholds (27). The same metric was applied to three reference GTs used to contextualise MOSAIC's performance: pairwise agreement between the three ground-truth frameworks (GT inter-agreement); agreement between the two independent assessor agents within each pipeline (inter-assessor agreement); and agreement between each assessor and the consolidated output, the latter two quantifying the contribution of the consolidation step. The structure of disagreement, adjacent- versus non-adjacent-tier, was examined using row-normalised confusion matrices for all pairwise comparisons.

*2.6.3 Third primary outcome - Predictive validity*

The predictive validity of the MOSAIC severity tiers for downstream outcomes was evaluated over the five-year follow-up window from the primary index date $T_5$ landmark. For all-cause mortality, survival was estimated using the Kaplan–Meier method and compared across severity tiers with the multivariate log-rank test, cumulative hazard was displayed using the Nelson–Aalen estimator, and restricted mean survival time (RMST) was computed at the five-year horizon (reported in days). Severity tier-specific hazards relative to the Baseline tier were estimated using Cox proportional-hazards models (with a small ridge penalty, penalizer = 0.01), reported both crude and adjusted for age and sex, as hazard ratios (HRs) with 95% CIs. For new diabetes complications, which are subject to the competing risk of death, cumulative incidence was estimated using the Aalen–Johansen estimator with death treated as a competing event, and five-year cumulative incidence is reported with 95% CIs. Tiers were compared using the multivariate log-rank test on the cause-specific event indicator.

*2.6.4 Sensitivity analyses*

Four sets of sensitivity analyses were performed. First, the incremental contribution of multi-agent reasoning over fixed rule execution was quantified by comparing Type 1 OW MOSAIC with Type 2 OW $\text{MOSAIC}_{\text{Frozen}}$ using both the population-level posterior tier-probability differences and patient-level κw. Second, CW versus OW agreement (κw) served as a sensitivity analysis for model-family dependence. Third, agreement (κw) between the two independent assessor agents within OW and CW (inter-assessor agreement); and agreement between each assessor and the consolidated output, to determine the contribution of the consolidation step. Fourth, representative cases were reviewed qualitatively by human experts to illustrate recurring disagreement mechanisms.

*2.6.5 Software and environment*

All analyses were performed in Python 3 on Google Colab Pro with an NVIDIA T4 GPU runtime. Survival and competing-risks analyses (Kaplan–Meier, Nelson–Aalen, Aalen–Johansen, Cox proportional-hazards, and log-rank tests) used lifelines 0.3.0.3; agreement metrics (quadratic-weighted Cohen's κ and confusion matrices) used scikit-learn 1.8.0; the Bayesian Dirichlet–multinomial model was implemented in PyMC 5.28.5 on the PyTensor backend, with posterior summaries (highest-density intervals, $\hat{R}$, and effective sample size) computed in ArviZ 0.22.0; exact Poisson confidence intervals for incidence rates and kernel density estimates for posterior visualisation used SciPy 1.16.3; and figures were produced with Matplotlib 3.10.0, seaborn 0.13.2, and python-ternary 1.0.8. All stochastic procedures (MCMC sampling and bootstrap resampling) were run under a fixed random seed (42) to ensure reproducibility.

### *2.7 Reporting Standards, Privacy, & Ethics*

The study was prepared in accordance with Tripod-LLM and Chart checklists (Supplementary Tables 7 and 8). All statistical analysis code, GT implementations, MOSAIC implementations, LLM prompt strategies, MOSAIC classification results, and statistical assessments are publicly available via a GitHub repository. The use of fully synthetic data eliminates the legal and governance barriers associated with real EHR access while enabling reproducible, large-scale evaluation, even when proprietary LLMs are used.

## *3. Results*

### *3.1 Framework Generation*

The Phase 1 pipeline retrieved three clinical sources for generating $MOSAIC_F$: 1) the ISPAD Clinical Practice Consensus Guidelines 2024 (28), 2) the IDF Clinical Practice Recommendations for Managing T2D in Primary Care (29), and 3) the Novo Nordisk Guideline Directed Management of Diabetes Comorbidities (30). The resulting framework operationalised T2D severity into four tiers (i.e., Baseline, Mild, Moderate, and Critical) across eight clinical domains: four biomarker domains (i.e., BMI in paediatric patients, HbA1c, fasting plasma glucose [FPG], and postprandial glucose [PPG]), comorbidities, residual beta-cell function, cardiorenal complications, and social determinants of health (SDOH). Severity was assigned by the highest tier triggered across domains, with mandatory escalation to "Critical" when two or more domains concurrently reached the "Moderate" tier or specific end-organ thresholds were reached; full tier-specific criteria for MOSAIC and all three GT frameworks are shown in Table 1.

### *3.2 Data Suitability*

Laboratory and vital-sign variables required across all frameworks (HbA1c, eGFR, UACR) were available for 83.4% of patients. Missing laboratory analyses or conditions do not represent missing data. Accordingly, condition domains showed prevalence-consistent representation: neuropathy (~65%) and retinopathy (~60%) were well-represented, while foot ulcer/amputation (1.5%) and peripheral vascular disease (<1%) were sparse, reflecting low clinical prevalence rather than data quality failure (Supplementary Figure 1).

### *3.3 Study population*

Of 1,593,365 synthetic patients, 35,368 met eligibility criteria at $T_5$. The closed-weight (CW) analytic cohort comprised 200 patients and the open-weight (OW) cohort 4,886 patients; both were broadly representative of the source population in sex, race, and age at T2D diagnosis. Metformin was the dominant first-line treatment across all groups (85–89%) (Supplementary Figure 2).

### *3.4 Second Primary Outcome - Classification Agreement*

#### *3.4.1 Population level*

Severity tier distributions and posterior tier probabilities (π) for the OW cohort are summarised in Table 2 (Panel A), with posterior distributions shown in Supplementary Figures 3–4 and Supplementary Table 9.

*Type 1*. Open-weight MOSAIC distributed patients across all four tiers (Baseline 24.0%, π = 0.240 [0.228–0.252]; Mild 25.3%, π = 0.255 [0.243–0.267]; Moderate 36.0%, π = 0.360 [0.347–0.374]; Critical 14.5%, π = 0.145 [0.135–0.154]). It assigned substantially fewer patients to Baseline than all three external GTs (40.3–44.1%), with every Baseline tier-probability difference (Δ = GT − OW MOSAIC) showing a 95% HDI excluding zero, and correspondingly more patients to the intermediate and advanced tiers. The pattern relative to each GT differed: against Young GT, OW MOSAIC assigned more patients to Moderate (36.0% vs 25.8%) and Critical (14.5% vs 4.1%); against Cooper GT, Moderate assignment was comparable (36.0% vs 38.0%, HDI including zero) but OW MOSAIC assigned fewer to Critical (14.5% vs 18.1%); and DiSSCo GT diverged most, concentrating patients in Mild (53.1%) and assigning none to Critical (0.0%).

*Type 2*. $MOSAIC_{Frozen}$ produced a markedly more conservative distribution, assigning 68.1% of patients to Baseline (π = 0.681), 3.1% to Mild, 27.4% to Moderate, and 1.6% to Critical, with each tier-probability difference from OW MOSAIC excluding zero. The Baseline contrast was the most pronounced of any comparison (posterior Δ Baseline = +0.441 [+0.423, +0.461]): the deterministic rule-based execution of $MOSAIC_F$ retained more than two-thirds of patients at Baseline, whereas the agentic pipeline redistributed a substantial share into higher-severity tiers. This confirms that multi-agent reasoning produced population-level classification behaviour well beyond fixed execution of the same framework (Supplementary Figure 3).

*CW vs OW*. Posterior tier probabilities for the closed-weight cohort (N = 200) were estimated separately and are shown alongside the matched open-weight subsample (same 200 patients) in Supplementary Figure 4 (closed-weight: Baseline 28.5%, Mild 21.0%, Moderate 39.0%, Critical 11.5%; open-weight matched subsample: Baseline 21.0%, Mild 15.0%, Moderate 51.0%, Critical 13.0%). Because similar marginal tier proportions do not guarantee agreement at the individual patient level, the two pipelines are additionally compared directly at the individual level (Section 3.4.2) rather than through marginal tier distributions alone; qualitatively, both pipelines produced the same low-Baseline, intermediate-weighted distribution that distinguished the agentic configuration from both $MOSAIC_{Frozen}$ and the external GTs.

### *3.4.2 Individual level*

Patient-level concordance, quantified using quadratic-weighted Cohen's κw, is summarised in Table 2 (Panel B); row-normalised confusion matrices for all pairwise comparisons are provided in Supplementary Figure 5.

*Type 1*. In the OW cohort, MOSAIC achieved moderate agreement with Cooper GT (κw = 0.597 [0.577–0.615]) and Young GT (κw = 0.534 [0.513–0.554]), and fair agreement with DiSSCo GT (κw = 0.320 [0.303–0.337]); neither the minimum-acceptable (κw ≥ 0.61) nor the target (κw ≥ 0.80) threshold was met against any external GT. In the closed-weight benchmark cohort, agreement was higher and met the minimum-acceptable threshold against both Cooper GT (κw = 0.702 [0.621–0.775]) and Young GT (κw = 0.671 [0.572–0.753]). GT inter-agreement provides essential context: Cooper GT and Young GT agreed almost perfectly (κw = 0.819 [0.812–0.826]), whereas DiSSCo GT fell below the moderate threshold against both (κw = 0.431 and 0.585), indicating that the lower MOSAIC–DiSSCo agreement reflects definitional divergence between severity frameworks rather than MOSAIC error alone. Disagreement concentrated at adjacent tier boundaries rather than extreme misclassification, with highest concordance for Baseline patients (96–97% across all GTs).

*Type 2*. MOSAICFrozen showed only moderate patient-level agreement with OW MOSAIC (κw = 0.428 [0.410–0.447]). Together with the substantial Baseline redistribution described in Section 3.4.1, this confirms that the agentic pipeline and the deterministic execution of the same underlying framework (MOSAICF) classified a meaningful proportion of patients differently, quantifying the contribution of multi-agent reasoning over fixed rule execution.

CW vs OW. Direct agreement between the closed- and open-weight pipelines was substantial (κw = 0.773 [0.700–0.829]) in the shared benchmark cohort, supporting the sensitivity analysis and indicating that the open-weight pipeline substantially reproduced closed-weight classification behaviour. The two pipelines nonetheless showed markedly different internal consolidation dynamics. In the CW pipeline, inter-assessor agreement (GPT-4o vs DeepSeek-V3) was substantial (κw = 0.621 [0.524–0.706]) and rose to almost perfect after consolidation (assessor-to-consolidated κw = 0.831–0.834), indicating that the consolidator performed substantive adjudication. In the OW pipeline, inter-assessor agreement (Gemma 2 vs Qwen 2.5) was already almost perfect (κw = 0.953 [0.949–0.957]), with similarly high assessor-to-consolidated agreement (κw = 0.911–0.959), suggesting the consolidator primarily stabilised rather than resolved disagreement.

### *3.5 Predictive Validity*

Over the five-year follow-up from the $T_5$ landmark, the OW cohort (N = 4,886) accrued 489 deaths over 22,820 person-years and 3,146 new diabetes complications over 18,290 person-years (Supplementary Tables 10 and 11). The severity gradient was clinically coherent across tiers: retinopathy prevalence rose from 1.6% in Baseline to 95.5% in Critical, nephropathy from 0.3% to 86.4%, and first-line insulin use from 2.1% to 29.3% (Supplementary Tables 10 and 11). Tier-specific incidence rates, restricted mean survival time (RMST), and Cox estimates are reported in Table 3; survival and cumulative-incidence curves are shown in Figure 2 (OW MOSAIC and Cooper GT) and Supplementary Figures 6-9.

#### *3.5.1 Overall mortality*

Type 1. Open-weight MOSAIC severity tiers produced significant survival separation (log-rank $p < 0.001$; Figure 2). Five-year Kaplan–Meier survival declined from 0.959 [0.946–0.969] in Baseline to 0.904 [0.886–0.920] in Mild, 0.863 [0.845–0.878] in Moderate, and 0.865 [0.838–0.889] in Critical, with a corresponding fall in RMST (1,799, 1,760, 1,704, and 1,708 days, respectively; Table 3) and rising crude mortality rates (0.83, 1.97, 2.93, and 2.87 per 100 person-years; Supplementary Table 11). The gradient was not strictly monotonic: the Moderate and Critical tiers showed closely overlapping survival, while both were clearly separated from Baseline and Mild. Crude Cox models showed significantly elevated hazards for every non-Baseline tier (Mild HR = 1.63 [1.26–2.11]; Moderate HR = 2.43 [1.93–3.07]; Critical HR = 2.34 [1.77–3.09]; all $p < 0.001$; Table 3), the overlapping Moderate and Critical estimates being consistent with these tiers representing a shared high-risk population. The three GT frameworks similarly stratified mortality (all log-rank $p < 0.001$; Supplementary Figure 9) but without a consistent monotonic gradient: Young GT showed its largest crude hazard in the Mild tier (HR = 3.16 [2.58–3.88]), Cooper GT in the Moderate tier (HR = 3.38 [2.77–4.12]) with a non-significant Critical tier (HR = 1.06 [0.79–1.43], $p = 0.694$), and DiSSCo GT the most progressive pattern (Mild HR = 2.47 [2.02–3.02]; Moderate HR = 4.03 [2.95–5.51]). MOSAIC thus reproduced the core mortality-stratification behaviour of established frameworks, and the non-monotonicity at its upper tiers mirrored that seen across the GTs.

*Type 2.* MOSAIC Frozen tiers also stratified mortality in the open-weight cohort (log-rank $p < 0.001$; Supplementary Figure 8). Although the frozen rule-based classifier assigned the majority of patients to Baseline and only 1.6% to Critical (Section 3.4.1), so that the Critical tier was small (N = 76), the higher tiers nonetheless contained sufficient events for stable estimation. Five-year Kaplan–Meier survival fell from 0.929 [0.919–0.937] in Baseline to 0.816 [0.752–0.865] in Mild, 0.855 [0.835–0.873] in Moderate, and 0.472 [0.353–0.582] in Critical, and all three non-Baseline tiers showed significantly elevated crude hazards (all $p < 0.001$; Table 3). The Critical tier carried by far the highest risk (HR = 9.79 [6.93–13.81]), reflecting a very high event rate (38 deaths among 76 patients) and the shortest RMST of any frozen tier (1,303 days, versus 1,655 in Mild and 1,705 in Moderate). As with the agentic open-weight pipeline the gradient was non-monotonic, but here the irregularity occurred at the lower tiers: Mild showed both lower five-year survival and a higher crude hazard than Moderate (HR = 2.63 [1.86–3.72] vs 1.97 [1.64–2.37]) before the sharp escalation at Critical.

*OW vs CW.* In the closed-weight benchmark cohort (N = 200), mortality was examined as a sensitivity analysis (Supplementary Figure Y). Both the closed-weight and open-weight pipelines separated mortality significantly despite the small sample (both log-rank $p < 0.001$) and in the same direction. Under the closed-weight pipeline, five-year survival fell monotonically from 1.000 in Baseline to 0.896 [0.745–0.960] in Mild, 0.660 [0.536–0.759] in Moderate, and 0.255 [0.084–0.470] in Critical, with correspondingly rising crude hazards (Mild HR = 2.61 [0.64–10.70], $p = 0.183$; Moderate HR = 10.93 [3.43–34.87], $p < 0.001$; Critical HR = 28.75 [8.49–97.43], $p < 0.001$; Table 3). Applying the open-weight pipeline to the same 200 patients reproduced the identical ordering (five-year survival 0.930, 0.716, and 0.411 across Mild, Moderate, and Critical; Moderate HR = 7.55 [2.22–25.68]; Critical HR =

18.94 [5.26–68.19]). As expected from the small tier sizes the confidence intervals were wide, and the absolute hazard magnitudes are not directly comparable with the full open-weight cohort owing to the different case-mix; nevertheless the consistent monotonic separation under both weight settings confirms that MOSAIC classification stratified mortality robustly even at this reduced scale.

### *3.5.2 New onset of T2D complications*

*Type 1.* New diabetes complications showed a consistent inverse gradient across OW MOSAIC tiers (log-rank $p < 0.001$; Figure 2): five-year cumulative incidence, with death treated as a competing risk, was highest in Baseline (0.705 [0.679–0.731]) and declined through Mild (0.547), Moderate (0.460), and Critical (0.308), mirrored by crude complication rates of 25.04, 18.97, 14.53, and 9.32 per 100 person-years, respectively (Supplementary Table 11). This inverse pattern is clinically interpretable as depletion of susceptibles, higher-severity patients have largely accrued, before the primary index date, the complications counted as incident events in lower-severity patients, and was reproduced across all three GT frameworks. Emergency encounters, by contrast, showed a non-monotonic pattern that peaked in the Mild tier (five-year cumulative incidence 0.277 [0.252–0.302]), with the peak tier varying by framework, suggesting these patterns reflect how each framework populates its tiers rather than a consistent severity–utilisation relationship.

*Type 2.* Five-year cumulative incidence of new diabetes complications differed significantly across MOSAIC$_{Frozen}$ tiers (log-rank $p < 0.001$), but the pattern was not strictly monotonic: incidence was highest and closely comparable in Baseline (0.609 [0.592–0.625]) and Mild (0.607 [0.536–0.678]), dropping sharply in Moderate (0.291 [0.266–0.316]) and remaining similarly low in Critical (0.311 [0.202–0.419]). As with mortality, the small higher-severity tiers under MOSAICFrozen (Mild n = 247; Critical n = 76) limit precision, reflected in Critical's markedly wider confidence interval.

## ***3.6 Representative examples of agreement and disagreement between MOSAIC and GTs***

Qualitative review by human experts of classifications between MOSAIC and GTs showed three recurring disagreement mechanisms: *insulin-dependent classification*, in which first-line insulin use triggered MOSAIC escalation that is not operationalised in any GT; *medication reason-code inference*, in which MOSAIC inferred cardiovascular disease from pharmacotherapy context unavailable to condition-code-only algorithms; and *two-domain* *concurrence*, in which two simultaneous Tier 3 signals triggered escalation to Tier 4 at a threshold absent from all comparator frameworks. Representative examples are provided in Supplementary Table 12.

## *4. Discussion*

MOSAIC successfully generated and applied a clinically coherent, literature-derived T2D severity framework, demonstrating the feasibility of agentic LLM systems for multidimensional phenotyping from longitudinal EHR data. The findings support a layered interpretation: agentic synthesis produced a richer framework than any single algorithmic comparator; multi-agent reasoning contributed classification behaviour beyond deterministic rule execution; severity tiers showed clinically meaningful associations with downstream mortality; and disagreements with ground truths were systematic rather than arbitrary, driven by identifiable differences in severity operationalisation.

### *4. 1 Framework generation and evidence synthesis*

The Phase 1 pipeline produced a framework extending substantially beyond the domain coverage of the three algorithmic comparators, incorporating glycaemic staging, quantitative beta-cell function markers (HOMA-IR, C-peptide, and medication escalation as surrogate indicators where direct biomarker data were unavailable), and social determinants of health alongside the complication-burden domains that anchor established frameworks. The consolidator agent resolved three inter-source discrepancies into operationalised tier logic, producing a classification-ready rubric rather than a narrative summary. These findings align with emerging evidence that LLMs can accelerate phenotype drafting and concept identification (31, 32) , but with an important caveat: hallucinated codes, incorrect thresholds, and incomplete inclusion criteria remain documented limitations of LLM-generated phenotype definitions, and the retrieval cap (max_results = 3) was selected based on RAG evidence that a limited set of high-quality sources outperforms larger, noisier retrieval sets (20, 31, 32), though as no alternative values were evaluated, this cannot be assumed to represent the optimal setting. Framework generation should therefore be understood as LLM-assisted drafting requiring expert validation, not autonomous guideline development. The contribution of the agentic pipeline is to accelerate and standardise the synthesis of dispersed evidence into a computable rubric; accountability for the resulting definitions, however, must remain with human experts. How safely this is achieved depends less on the underlying model than on how the human–AI interaction is structured, and several features appear essential to an effective human-in-the-loop design. First, *source governance*: the literature retrieved for synthesis should be curated or pre-approved by domain experts rather than drawn from unconstrained web search, and each criterion in the generated framework should carry explicit provenance to its source, allowing reviewers to audit the evidence behind every threshold rather than the artefact as a whole. Second, *staged review*: oversight is most effective when applied at intermediate checkpoints, source selection, extracted criteria, numeric cut-points, and escalation logic, rather than to the final output alone, since an inappropriate threshold or mis-transcribed cut-off introduced early is difficult to detect downstream. Third, *conflict surfacing*: where source guidelines diverge, the system should expose the discrepancy and its candidate resolutions for expert adjudication rather than resolving it silently, preserving clinical authority over contested definitions. Fourth, *operational and face validity*: each criterion should be confirmed both clinically sensible and computable from the available data and vocabularies, a check that would flag, for instance, a criterion inherited from a source guideline that is inapplicable to the target population, or a laboratory value not routinely recorded in the EHR. Finally, *versioned sign-off*: expert modifications, their rationale, and a named clinical authorisation should be recorded, giving the framework an auditable revision history and a clear line of accountability before patient-level application. Structured in this way, the pipeline amplifies rather than replaces expert effort, compressing the initial draft while keeping every definition traceable, with human judgement governing the decisions that determine clinical validity. This design also offers a practical route to maintaining severity definitions as guidelines evolve: re-running the synthesis over updated sources yields a revised draft for expert re-validation, rather than requiring de novo manual redevelopment.

#### *4. 2 Classification agreement and agentic contribution*

OW MOSAIC achieved moderate agreement with Cooper GT (κw = 0.597) and Young GT (κw = 0.534), and fair agreement with DiSSCo GT (κw = 0.320), not meeting the pre-specified minimum acceptable threshold against any external comparator; the CW pipeline met this threshold against Cooper GT (κw = 0.702) and Young GT (κw = 0.671), confirming the criterion is achievable under stronger model configurations. A central interpretive challenge is that the three GT frameworks are operational comparators, not absolute clinical ground truth. Sood et al. showed that five established computable T2D definitions shared only 22.7% of patients (33), and the present results confirm this: Cooper and Young GT reached near-perfect inter-agreement (κw = 0.819), while DiSSCo diverged from both below the minimum acceptable threshold, meaning an established validated framework failed the same standard applied to MOSAIC in this context. Disagreement with a single GT should therefore be interpreted as operationalisation-specific alignment rather than error per se. The more substantive finding was the comparison with MOSAIC Frozen: substantial patient redistribution from Baseline toward intermediate and advanced tiers, combined with only moderate patient-level agreement (κw = 0.428), confirms that multi-agent reasoning captured clinical signals beyond deterministic threshold logic (34, 35). The closed-weight pipeline's stronger GT agreement and consolidator adjudicative role (κw rising from 0.621 to >0.83 post-consolidation) support proprietary models for higher-stakes tasks, while the open-weight pipeline's substantial reproduction of closed-weight behaviour (κw = 0.773) supports local deployment where data governance requires it (36-39).

### *4. 3 Predictive validity*

MOSAIC tiers showed the most interpretable associations with all-cause mortality, reproducing the core validation pattern of established frameworks, progressive survival separation, increasing cumulative hazard, and significant Cox hazard ratios, consistent with mortality gradients reported in original DCSI, DiSSCo, and Cooper GT validation studies. Non-strict monotonicity between Moderate and Critical tiers was not unique to MOSAIC: Young GT showed its largest hazard in Mild and Cooper GT in Moderate, indicating that non-monotonicity at upper tier boundaries reflects tier-population composition rather than classification failure. The inverse complications gradient, highest cumulative incidence in Baseline, lowest in Critical, is clinically interpretable as depletion of susceptibles: higher-severity patients have largely accumulated complications prior to the index date, leaving fewer incident events to observe prospectively (40-43). Emergency encounters showed non-monotonic separation consistent with the known sensitivity of all-cause ED utilisation to care-seeking behaviour, coding variation, and access factors beyond biological severity (41).

### *4. 4 Evidence transparency*

Disagreements were structured around three recurring mechanisms: treatment intensity as a severity signal absent from Young and DiSSCo GT; medication-linked surrogate diagnoses extending beyond diagnosis-code logic; and multi-domain concurrence escalation via Rule 0a. Medical history and complications remained the shared evidential backbone across all frameworks, confirming that MOSAIC extended rather than replaced conventional severity markers. A notable finding was that closed-weight models incorporated sociodemographic information in substantially more cases than open-weight models, consistent with evidence that model families differ in sensitivity to contextual cues (44, 45). While the clinical implications of this difference remain unclear in the present context, the differential weighting of social determinants across configurations is worth monitoring in future work, given documented evidence that clinical LLMs can produce differential outputs based on sociodemographic labels (46, 47).

### *4.1 Strengths and Limitations*

Key strengths include the end-to-end design spanning framework generation, classification, agreement analysis, predictive validity, and qualitative reasoning review; validation against three conceptually distinct GT frameworks, allowing GT inter-agreement itself to contextualise MOSAIC's performance; and the inclusion of MOSAIC Frozen as an internal comparator enabling explicit quantification of the agentic reasoning contribution. The primary limitation is the use of synthetic rather than real-world EHR data: while SyntheticMass enabled privacy-preserving, reproducible experimentation, prior work has shown it may not accurately reproduce real-world clinical heterogeneity, missingness, or deviation from guideline-based care (48), and findings should be interpreted as proof-of-concept evidence rather than real-world performance estimates. Finally, prompts and orchestration were not exhaustively optimised, and the very high open-weight inter-assessor agreement raises the question of whether the full multi-agent architecture was necessary in that configuration (49, 50).

## ***5. Conclusion***

MOSAIC demonstrates that agentic LLM systems can derive and apply clinically meaningful T2D severity phenotypes from structured longitudinal EHR data, producing tiers that overlap with established frameworks, contribute reasoning beyond deterministic rule execution, and stratify downstream mortality risk. Disagreements with algorithmic comparators reflect identifiable differences in severity operationalisation rather than arbitrary error, and the findings highlight that disease severity is a complex construct shaped by the clinical domains and logic used to define it. MOSAIC should be interpreted as a proof-of-concept rather than a deployable clinical system. Notably, the open-weight pipeline substantially reproduced closed-weight classification behaviour at lower agreement with the external frameworks, indicating that locally deployable open-source models are a credible alternative to proprietary APIs, a consideration that matters not only for the per-query costs and usage limits of commercial models at population scale, but also for the data-governance constraints of processing patient records through external services. Future work should prioritise validation on real-world EHR data, expert adjudication of classifications, formal fairness evaluation, and extension to chronic diseases where severity is similarly multidimensional.

### *6. Ethics, Funding, and Conflict of Interest*

No funding was obtained for this study. Because the study relied exclusively on publicly available synthethic data, ethical approval and informed consent were not applicable. The authors declare no conflicts of interest with respect to the conduct or reporting of this study.

### *7. Authors' Contribution*

MDCS, AQV, NSH, DR, and MS had primary responsibility for the study conception and design and for the acquisition, analysis, and interpretation of data, and they led the drafting of the manuscript. MDCS, AQV, and MS contributed to data collection, formal analysis, interpretation of results, and revision of the manuscript. All authors read and approved the final manuscript and agree to be accountable for the work, including ensuring that any questions related to the accuracy or integrity of the study are appropriately addressed.

### *8. Acknowledgement*

The authors thank Debora Graf for her careful reading of the manuscript and for her valuable comments and revisions, which substantially improved the clarity and quality of this work.

*9. References*

1. Banda JM, Seneviratne M, Hernandez-Boussard T, Shah NH. Advances in electronic phenotyping: from rule-based definitions to machine learning models. Annual review of biomedical data science. 2018;1(1):53–68.
2. Pendergrass SA, Crawford DC. Using electronic health records to generate phenotypes for research. Current protocols in human genetics. 2019;100(1):e80.
3. Mohammadi S, Campbell C, Sturkenboom MC, Vaz TA. A Systematic Review to Summarize and Critically Appraise Existing Phenotype Libraries Using Electronic Health Records. Pharmacoepidemiol Drug Saf. 2026;35(5):e70378.
4. Yang S, Varghese P, Stephenson E, Tu K, Gronsbell J. Machine learning approaches for electronic health records phenotyping: a methodical review. Journal of the American Medical Informatics Association. 2023;30(2):367–381.
5. Shivade C, Raghavan P, Fosler-Lussier E, Embi PJ, Elhadad N, Johnson SB, Lai AM. A review of approaches to identifying patient phenotype cohorts using electronic health records. Journal of the American Medical Informatics Association. 2014;21(2):221–230.
6. Cooper J, Jackson T, Haroon S, Crowe FL, Hathaway E, Fitzsimmons L, Nirantharakumar K. Defining phenotypes of disease severity for long-term cardiovascular, renal, metabolic, and mental health conditions in primary care electronic health records: A mixed-methods study using the nominal group technique. Journal of Biomedical Informatics. 2025;166:104831.
7. Reddy K, Sinha P, O'Kane CM, Gordon AC, Calfee CS, McAuley DF. Subphenotypes in critical care: translation into clinical practice. The Lancet Respiratory Medicine. 2020;8(6):631–643.
8. Nair ATN, Wesolowska-Andersen A, Brorsson C, Rajendrakumar AL, Hapca S, Gan S, Dawed AY, Donnelly LA, McCrimmon R, Doney AS. Heterogeneity in phenotype, disease progression and drug response in type 2 diabetes. Nature medicine. 2022;28(5):982–988.
9. Gordon AC, Alipanah-Lechner N, Bos LD, Dianti J, Diaz JV, Finfer S, Fujii T, Giamarellos-Bourboulis EJ, Goligher EC, Gong MN. From ICU syndromes to ICU subphenotypes: consensus report and recommendations for developing precision medicine in the ICU. American journal of respiratory and critical care medicine. 2024;210(2):155–166.
10. Wang L, Chen X, Deng X, Wen H, You M, Liu W, Li Q, Li J. Prompt engineering in consistency and reliability with the evidence-based guideline for LLMs. NPJ digital medicine. 2024;7(1):41.
11. Walonoski J, Kramer M, Nichols J, Quina A, Moesel C, Hall D, Duffett C, Dube K, Gallagher T, McLachlan S. Synthea: An approach, method, and software mechanism for generating synthetic patients and the synthetic electronic health care record. Journal of the American Medical Informatics Association. 2018;25(3):230–238.
12. Duchesneau ED, Jackson BE, Webster-Clark M, Lund JL, Reeder-Hayes KE, Nápoles AM, Strassle PD. The timing, the treatment, the question: comparison of epidemiologic approaches to minimize immortal time bias in real-world data using a surgical oncology example. Cancer Epidemiology, Biomarkers & Prevention. 2022;31(11):2079–2086.
13. Carroll OU, Bidulka P, Basu A, Adler AI, O'Neill S, Briggs AH, Lugo-Palacios DG, Khunti K, Grieve R. Long-term outcomes following alternative second-line oral glucose-lowering treatments: Results from the real-world progression in type 2 diabetes mellitus United Kingdom (RAPIDS-UK) model. Diabetes, Obesity and Metabolism. 2025;27(8):4181–4191.
14. Lee Y-B, Han K, Kim B, Park SH, Hur KY, Kim G, Kim JH, Jin S-M. Time to Insulin Therapy and Severe Hypoglycemia in Korean Adults Initially Diagnosed with Type 2 Diabetes: A Nationwide Study. Endocrinology and Metabolism. 2025;40(3):421–433.
15. Vajravelu ME, Hitt TA, Amaral S, Levitt Katz LE, Lee JM, Kelly A. Real-world treatment escalation from metformin monotherapy in youth-onset Type 2 diabetes mellitus: A retrospective cohort study. Pediatric diabetes. 2021;22(6):861–871.
16. Huo L, Magliano DJ, Rancière F, Harding JL, Nanayakkara N, Shaw JE, Carstensen B. Impact of age at diagnosis and duration of type 2 diabetes on mortality in Australia 1997–2011. Diabetologia. 2018;61(5):1055–1063.
17. Vittinghoff E, McCulloch CE. Relaxing the rule of ten events per variable in logistic and Cox regression. American journal of epidemiology. 2007;165(6):710–718.
18. Duan Z, Wang J. Exploration of llm multi-agent application implementation based on langgraph+ crewai. arXiv preprint arXiv:241118241. 2024.
19. Rathod SG, Asawale S, Yashwante M, Taware RD. Beyond Text Matching: Multi-Agent System for Comprehensive Research Paper Evaluation. In.2025 IEEE Pune Section International Conference (PuneCon): IEEE; 2025. p. 1–6.
20. Levy S, Mazor N, Shalmon L, Hassid M, Stanovsky G. More documents, same length: Isolating the challenge of multiple documents in rag. arXiv preprint arXiv:250304388. 2025.
21. Liu S, McCoy AB, Wright A. Improving large language model applications in biomedicine with retrieval-augmented generation: a systematic review, meta-analysis, and clinical development guidelines. Journal of the American Medical Informatics Association. 2025;32(4):605–615.
22. Tian F, Ganguly D, Macdonald C. Is relevance propagated from retriever to generator in RAG? In.European Conference on Information Retrieval: Springer; 2025. p. 32–48.
23. Cheng Y, Zhang C, Zhang Z, Meng X, Hong S, Li W, Wang Z, Wang Z, Yin F, Zhao J. Exploring large language model based intelligent agents: Definitions, methods, and prospects. arXiv preprint arXiv:240103428. 2024.
24. Myrzakhan A, Bsharat SM, Shen Z. Open-llm-leaderboard: From multi-choice to open-style questions for llms evaluation, benchmark, and arena. arXiv preprint arXiv:240607545. 2024.
25. Young BA, Lin E, Von Korff M, Simon G, Ciechanowski P, Ludman EJ, Everson-Stewart S, Kinder L, Oliver M, Boyko EJ. Diabetes complications severity index and risk of mortality, hospitalization, and healthcare utilization. The American journal of managed care. 2008;14(1):15.
26. Zghebi SS, Mamas MA, Ashcroft DM, Salisbury C, Mallen CD, Chew-Graham CA, Reeves D, Van Marwijk H, Qureshi N, Weng S. Development and validation of the DIabetes Severity SCOre (DISSCO) in 139 626 individuals with type 2 diabetes: a retrospective cohort study. BMJ Open Diabetes Research & Care. 2020;8(1).
27. McHugh ML. Interrater reliability: the kappa statistic. Biochemia medica. 2012;22(3):276–282.
28. Shah AS, Barrientos-Pérez M, Chang N, Fu J-F, Hannon TS, Kelsey M, Peña AS, Pinhas-Hamiel O, Urakami T, Wicklow B. ISPAD clinical practice consensus guidelines 2024: type 2 diabetes in children and adolescents. Hormone Research in Paediatrics. 2025;97(6):555–583.
29. Aschner P. New IDF clinical practice recommendations for managing type 2 diabetes in primary care. Diabetes research and clinical practice. 2017;132:169–170.

30. Nordisk N. Guideline Directed Management of Diabetes Comorbidities. In.: Novo Nordisk Medical; 2024.
31. Yan C, Ong HH, Grabowska ME, Krantz MS, Su W-C, Dickson AL, Peterson JF, Feng Q, Roden DM, Stein CM. Large language models facilitate the generation of electronic health record phenotyping algorithms. Journal of the American Medical Informatics Association. 2024;31(9):1994–2001.
32. Tekumalla R, Banda JM. Towards automated phenotype definition extraction using large language models. Genomics & Informatics. 2024;22(1):21.
33. Sood PD, Liu S, Kalyani RR, Kharrazi H. Comparing computable type 2 diabetes phenotype definitions in identifying populations of interest for clinical research. BMJ Health & Care Informatics. 2025;32(1):e101378.
34. Prakash R, Dupre ME, Østbye T, Xu H. Extracting critical information from unstructured clinicians' notes data to identify dementia severity using a rule-based approach: feasibility study. JMIR aging. 2024;7(1):e57926.
35. Speyer CB, Li D, Guan H, Yoshida K, Stevens E, Jorge AM, Everett BM, Feldman CH, Costenbader KH. Comparison of an administrative algorithm for SLE disease severity to clinical SLE Disease Activity Index scores. Rheumatology international. 2020;40(2):257–261.
36. Riedemann L, Labonne M, Gilbert S. The path forward for large language models in medicine is open. npj Digital Medicine. 2024;7(1):339.
37. Sandmann S, Hegselmann S, Fujarski M, Bickmann L, Wild B, Eils R, Varghese J. Benchmark evaluation of DeepSeek large language models in clinical decision-making. Nature medicine. 2025;31(8):2546–2549.
38. Dennstädt F, Hastings J, Putora PM, Schmerder M, Cihoric N. Implementing large language models in healthcare while balancing control, collaboration, costs and security. NPJ digital medicine. 2025;8(1):143.
39. Zhang G, Jin Q, Zhou Y, Wang S, Idnay B, Luo Y, Park E, Nestor JG, Spotnitz ME, Soroush A. Closing the gap between open source and commercial large language models for medical evidence summarization. NPJ digital medicine. 2024;7(1):239.
40. Yola M, Lucien A. Evidence of the depletion of susceptibles effect in non-experimental pharmacoepidemiologic research. Journal of clinical epidemiology. 1994;47(7):731–737.
41. Calip GS, Yerram P, Ascha MS. Nonrandomized Comparison of Adverse Events Following Facility-and Home-Infused Biologics Using Real-World Data. JAMA Network Open. 2021;4(6):e2111156.
42. Cheng L-J, Chen J-H, Lin M-Y, Chen L-C, Lao C-H, Luh H, Hwang S-J. A competing risk analysis of sequential complication development in Asian type 2 diabetes mellitus patients. Scientific Reports. 2015;5(1):15687.
43. Kunina H, Franzén S, Kjellsson MC. Modeling of Disease Progression of Type 2 Diabetes Using Real-World Data: Quantifying Competing Risks of Morbidity and Mortality. CPT: Pharmacometrics & Systems Pharmacology. 2025;14(3):606–615.
44. Hager P, Jungmann F, Holland R, Bhagat K, Hubrecht I, Knauer M, Vielhauer J, Makowski M, Braren R, Kaissis G. Evaluation and mitigation of the limitations of large language models in clinical decision-making. Nature medicine. 2024;30(9):2613–2622.
45. Tordjman M, Liu Z, Yuce M, Fauveau V, Mei Y, Hadjadj J, Bolger I, Almansour H, Horst C, Parihar AS. Comparative benchmarking of the DeepSeek large language model on medical tasks and clinical reasoning. Nature medicine. 2025;31(8):2550–2555.

46. Omar M, Soffer S, Agbareia R, Bragazzi NL, Apakama DU, Horowitz CR, Charney AW, Freeman R, Kummer B, Glicksberg BS. Sociodemographic biases in medical decision making by large language models. Nature Medicine. 2025;31(6):1873–1881.
47. Zack T, Lehman E, Suzgun M, Rodriguez JA, Celi LA, Gichoya J, Jurafsky D, Szolovits P, Bates DW, Abdulnour R-EE. Assessing the potential of GPT-4 to perpetuate racial and gender biases in health care: a model evaluation study. The Lancet Digital Health. 2024;6(1):e12–e22.
48. Chen J, Chun D, Patel M, Chiang E, James J. The validity of synthetic clinical data: a validation study of a leading synthetic data generator (Synthea) using clinical quality measures. BMC medical informatics and decision making. 2019;19(1):44.
49. Liu Y, Carrero ZI, Jiang X, Ferber D, Wölflein G, Zhang L, Jayabalan S, Lenz T, Hui Z, Kather JN. Benchmarking large language model-based agent systems for clinical decision tasks. npj Digital Medicine. 2026.
50. Spieser J, Balapour A, Meller J, Patra KC, Shamsaei B. A review of multi-agent AI systems for biological and clinical data analysis. Methods and Protocols. 2026;9(2):33.

**Table 1.** Comparison of diabetes complication severity tiering methods.

| TIER | **Young GT (DCSI)**<br>Cumulative complication score · SNOMED-CT codes + labs | **DiSSCo GT**<br>Severity-weighted sum · SNOMED-CT + procedures + UACR lab | **Cooper GT**<br>Priority-ordered flag logic · SNOMED-CT + LOINC + RxNorm | **MOSAIC**<br>Highest-domain threshold + multi-domain escalation (Rule 0a) · eight clinical domains |
|---|---|---|---|---|
| **Baseline** (Tier 1) | **DCSI = 0**<br>No qualifying SNOMED-CT codes in any of seven domains; creatinine <1.5 mg/dL; UACR <30 mg/g. | **Score = 0**<br>No conditions or procedures matching the 28 implemented domains; UACR <30 mg/g. | **No flags**<br>No complication codes in any of eight categories; eGFR ≥60 or not recorded; UACR <30 mg/g or not recorded; HbA1c ≤8.0% or not recorded; ≤1 drug class detected. | **HbA1c:** 6.5–<7.0%<br>**FPG:** 100–125 mg/dL<br>**PPG:** 140–199 mg/dL<br>**BMI (Paediatric):** ≥85th percentile<br>**Comorbidities:** None significant<br>**Cardiorenal:** eGFR ≥60; no HF/ASCVD<br>**Beta-cell / Pharmacotherapy:** HOMA-IR <2.5; C-peptide normal; ≤1 OAD class (metformin monotherapy proxy)<br>**SDOH:** No escalation flags |
| **Mild** (Tier 2) | **DCSI = 1**<br>Exactly one domain-point from any single trigger: background retinopathy (ret=1), TIA (cbv=1), stable angina (cvd=1), PVD code (pvd=1), neuropathy code (neu=1), hypoglycaemia (met=1), diabetic nephropathy code (neph=1), UACR ≥30 mg/g (neph=1), or creatinine 1.5–2.0 mg/dL (neph=1). | **0 < Score ≤ 3**<br>Minimum entry: hypertension or dyslipidaemia alone (w=0.5 each). Examples: neuropathy (1), retinopathy (1), foot ulcer (1), TIA (2), TIA + hypertension (2.5), or TIA + neuropathy (3). | **HbA1c or medication flag**<br>No macro/microvascular flags and no Critical triggers, but: maximum HbA1c >8.0% or more than one distinct drug class detected (from Metformin, Insulin, Sulfonylurea). | **HbA1c:** 7.0–<8.0%<br>**FPG:** 126–154 mg/dL<br>**PPG:** 200–299 mg/dL<br>**BMI (Paediatric):** ≥95th percentile<br>**Comorbidities:** HTN or dyslipidaemia<br>**Cardiorenal:** eGFR 30–59 or NYHA I–II HF without ASCVD<br>**Beta-cell / Pharmacotherapy:** HOMA-IR ≥2.5 or C-peptide reduced; ≥2 OAD classes<br>**SDOH:** ≥2 SDOH risk factors present |
| **Moderate** (Tier 3) | **DCSI = 2–3**<br>Two to three domain-points in any combination: proliferative retinopathy (ret=2), stroke (cbv=2), MI or heart failure (cvd=2), DKA (met=2), or severe nephropathy code / creatinine >2.0 mg/dL (neph=2); or any two single-point domains combined. | **3 < Score ≤ 8**<br>Combinations of medium-to-high-weight domains: e.g. heart failure (3) + retinopathy (1) = 4; MI/ACS (2) + TIA (2) + neuropathy (1) = 5; cardiac arrest (3) + DKA (2) + hypertension (0.5) = 5.5. | **Macro or micro flag**<br>Any macrovascular complication code (stroke/CVB, coronary/MI, heart failure, PVD) or any microvascular indicator (nephropathy/CKD, retinopathy, or neuropathy codes; UACR ≥30 mg/g; or eGFR <60 mL/min). Critical criteria not met. | **HbA1c:** ≥8.0–<11.0%<br>**FPG:** >154 mg/dL<br>**PPG:** ≥300 mg/dL<br>**BMI (Paediatric):** >99th percentile<br>**Comorbidities:** CVD (CAD/stroke/HF/PVD) or HTN + dyslipidaemia<br>**Cardiorenal:** eGFR 15–29 or ASCVD + HF or UACR >300 mg/g |

| TIER | **Young GT (DCSI)** Cumulative complication score · SNOMED-CT codes + labs | **DiSSCo GT** Severity-weighted sum · SNOMED-CT + procedures + UACR lab | **Cooper GT** Priority-ordered flag logic · SNOMED-CT + LOINC + RxNorm | **MOSAIC** Highest-domain threshold + multi-domain escalation (Rule 0a) · eight clinical domains |
|---|---|---|---|---|
| | | | | **Beta-cell / Pharmacotherapy:** C-peptide <0.2 nmol/L or ≥2 OAD class failures; insulin or ≥3 OAD classes<br>**SDOH:** Mandatory review flag triggered |
| **Critical** (Tier 4) | **DCSI ≥ 4**<br>Accumulation across domains: e.g. stroke (2) + MI/HF (2) = 4; nephropathy (2) + retinopathy (2) = 4; any combination of domain-points summing to four or more. | **Score > 8**<br>High-weight event combinations required: e.g. stroke (4) + heart failure (3) + neuropathy (1) + hypertension (0.5) = 8.5; amputation (3) + ESRD (3) + TIA (2) + neuropathy (1) = 9. | **Priority rule — first match wins**<br>Any one of: (1) ≥3 distinct complication categories present (from eight); (2) foot ulcer or amputation code; (3) retinopathy stage 3 (proliferative or macular oedema); (4) minimum eGFR <30 mL/min. | *Rule 0a — escalation to Tier 4 when ≥2 concurrent Tier 3 domain criteria are met, or insulin <15, or insulin dependence + macrovascular disease ≥2 territories, or HbA1c ≥11% + 1 other Tier 3/4 domain.*<br>**HbA1c:** ≥11.0% + ≥1 other Tier 3/4 domain (Rule 0a)<br>**FPG:** >154 mg/dL + ≥1 other Tier 3/4 domain (Rule 0a)<br>**PPG:** ≥300 mg/dL + ≥1 other Tier 3/4 domain (Rule 0a)<br>**BMI (Paediatric):** >99th percentile + ≥1 other Tier 3/4 domain (Rule 0a)<br>**Comorbidities:** Macrovascular disease ≥2 vascular territories + insulin dependence<br>**Cardiorenal:** eGFR <15 or dialysis or CKD stage 4 + ASCVD + HF<br>**Beta-cell / Pharmacotherapy:** Insulin + ≥2 OAD failure + concurrent Tier 3/4 domain criterion (Rule 0a)<br>**SDOH:** Tier 3 SDOH flag + concurrent Tier 3 domain criterion (requires human adjudication) |

***Legend:*** *GT, ground truth; DCSI, Diabetes Complications Severity Index; SNOMED-CT, Systematized Nomenclature of Medicine Clinical Terms; LOINC, Logical Observation Identifiers Names and Codes; UACR, urine albumin-to-creatinine ratio; eGFR, estimated glomerular filtration rate; HbA1c, glycated haemoglobin; FPG, fasting plasma glucose; PPG, postprandial plasma glucose; HTN, hypertension; CVD, cardiovascular disease; HF, heart failure; ASCVD, atherosclerotic cardiovascular disease; PVD, peripheral vascular disease; DKA, diabetic ketoacidosis; CAD, coronary artery disease; TIA, transient ischaemic attack; CKD, chronic kidney disease; ESRD, end-stage renal disease; OAD, oral antidiabetic drug; HOMA-IR, homeostasis model assessment of insulin resistance; SDOH, social determinants of health; NYHA, New York Heart Association.*

*DiSSCo domain weights (this implementation): hypertension = 0.5; neuropathy, retinopathy, nephropathy, foot ulcer, hypoglycaemia, stable angina, afib, PVD, cardiomyopathy, CHD, heart valve = 1; gangrene, laser therapy, blindness, albuminuria, DKA, MI/ACS, endocarditis, coronary intervention, TIA = 2; amputation, ESRD, HHS, cardiac arrest, heart failure = 3; stroke = 4. Cooper GT tier assignment follows strict priority order: Critical is evaluated first and takes precedence over all lower tiers. MOSAIC tier assignment follows Rule 0 (highest domain tier wins); Tier 4 entries in glycaemic and beta-cell domains are conditional on Rule 0a only.*

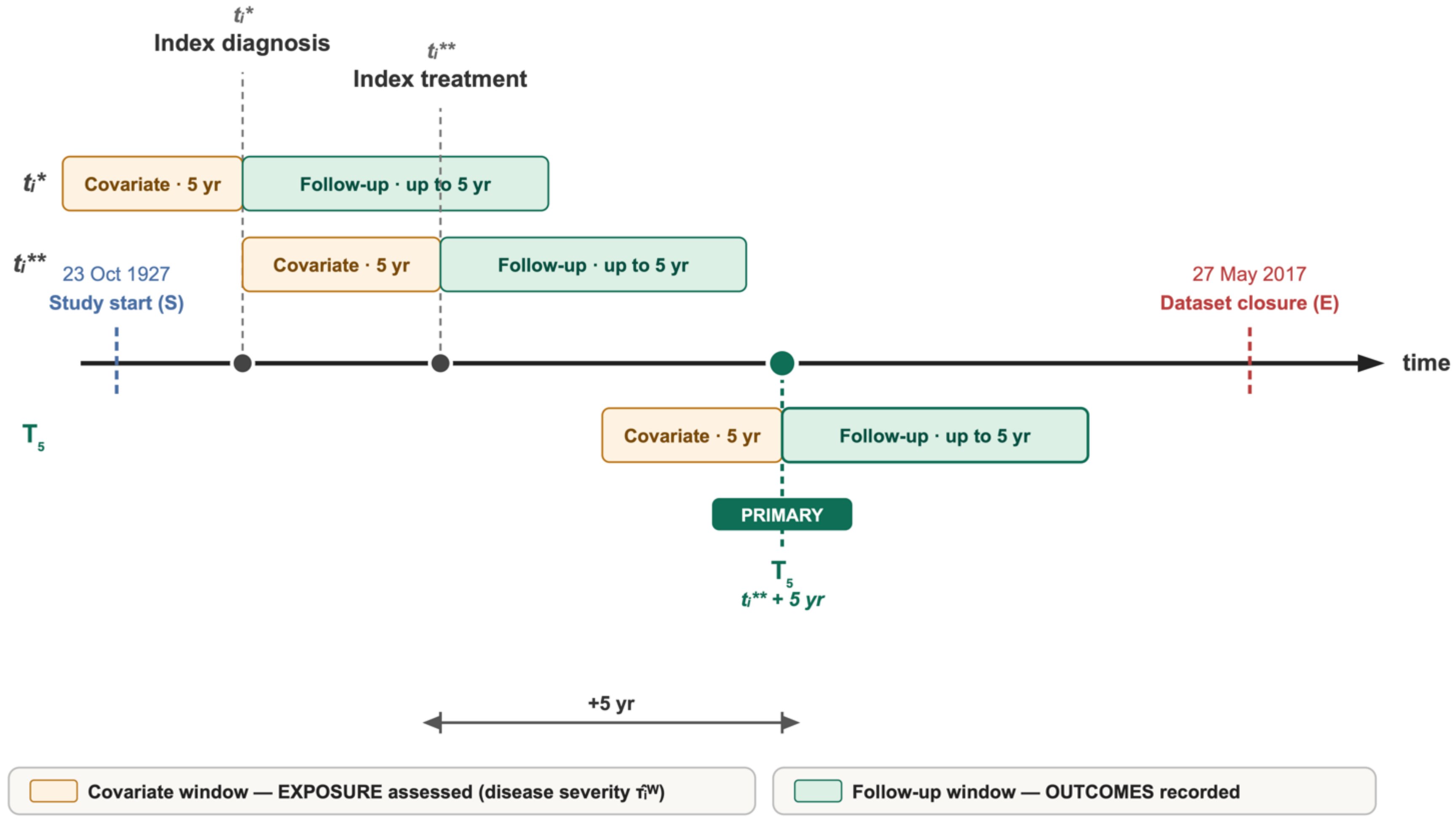

**Figure 1.** Multi-index date selection and observation window architecture

***Legend:*** *The study period is structurally bounded by the dataset [S, E] dates. Patients are tracked from initial type 2 diabetes diagnosis ($t_i^*$) through the initiation of first pharmacological treatment $t_i^{**}$. $T_5$ indicates the primary index date wich is five years after first treatment. Exposure is assessed in the covariate assessment window assessment, computing disease severity ($\hat{\tau}_i^W$), while outcomes are systematically monitored over standardized 5-year follow-up windows or terminating at death.*

**Table 2.** Tier distribution and patient-level agreement across classifiers.

**Panel A —** Severity tier assignment and posterior tier probabilities (OW cohort, N = 4,886)

| Tier | OW MOSAIC | Young GT | Cooper GT | DiSSCo GT | MOSAIC Frozen |
|---|---|---|---|---|---|
| **Baseline** | 1,172 (24.0%)<br><br>π = 0.240 [0.228–0.252] | 2,171 (44.1%)<br><br>π = 0.441† | 1,983 (40.3%)<br><br>π = 0.403† | 2,040 (41.5%)<br><br>π = 0.415† | 3,351 (68.1%)<br><br>π = 0.681† |
| **Mild** | 1,236 (25.3%)<br><br>π = 0.255 [0.243–0.267] | 1,278 (26.0%)<br><br>π = 0.260 | 186 (3.8%)<br><br>π = 0.038† | 2,615 (53.1%)<br><br>π = 0.531† | 150 (3.1%)<br><br>π = 0.031† |
| **Moderate** | 1,761 (36.0%)<br><br>π = 0.360 [0.347–0.374] | 1,267 (25.8%)<br><br>π = 0.258† | 1,862 (38.0%)<br><br>π = 0.379 | 263 (5.4%)<br><br>π = 0.054† | 1,341 (27.4%)<br><br>π = 0.273† |
| **Critical** | 709 (14.5%)<br><br>π = 0.145 [0.135–0.154] | 202 (4.1%)<br><br>π = 0.041† | 887 (18.1%)<br><br>π = 0.180† | 0 (0.0%)<br><br>π = 0.000† | 76 (1.6%)<br><br>π = 0.016† |

***Legend:*** *Each cell reports the number of patients assigned to that tier (n), the corresponding percentage of the cohort (%), and the posterior tier probability estimated from a Bayesian Dirichlet–Multinomial model (π). For OW MOSAIC, π is shown with its 95% highest density interval (HDI); for all comparator frameworks, π represents the posterior mean only. † indicates that the 95% HDI of the tier-probability difference (Δk = comparator − OW MOSAIC) excludes zero, confirming a systematic difference in how that tier is populated relative to OW MOSAIC.*

**Panel B** — Patient-level agreement (quadratic weighted Cohen's κw)

| Comparison | N | κw [95% CI] | Interpretation |
|---|---|---|---|
| **OW MOSAIC vs GT** | | | |
| vs Cooper GT | 4,886 | 0.597 [0.577–0.615] | Moderate |
| vs Young GT | 4,886 | 0.534 [0.513–0.554] | Moderate |
| vs DiSSCo GT | 4,886 | 0.320 [0.303–0.337] | Fair |
| vs MOSAIC Frozen | 4,886 | 0.428 [0.410–0.447] | Moderate |
| **CW MOSAIC vs GT** | | | |
| vs Cooper GT | 200 | 0.702 [0.621–0.775] | Substantial |
| vs Young GT | 200 | 0.671 [0.572–0.753] | Substantial |
| vs OW MOSAIC | 200 | 0.773 [0.700–0.829] | Substantial |
| **GT inter-agreement (reference)** | | | |
| Cooper GT vs Young GT | 4,886 | 0.819 [0.812–0.826] | Almost perfect |
| Cooper GT vs DiSSCo GT | 4,886 | 0.431 [0.416–0.446] | Moderate |
| Young GT vs DiSSCo GT | 4,886 | 0.585 [0.568–0.603] | Moderate |
| **Internal pipeline agreement** | | | |
| CW: inter-assessor (GPT-4o vs DeepSeek) | 200 | 0.621 [0.524–0.706] | Substantial |
| CW: assessors vs consolidated output (A1, A2) | 200 | 0.831–0.834 | Almost perfect |
| OW: inter-assessor (Gemma vs Qwen) | 4,769 | 0.953 [0.949–0.957] | Almost perfect |
| OW: assessors vs consolidated output (A1, A2) | 4,770–4,791 | 0.911–0.959 | Almost perfect |

***Legend:*** *Each row reports patient-level agreement between two classifiers quantified using quadratic weighted Cohen's κw, which penalises larger tier disagreements more heavily than adjacent-tier mismatches. κw values are shown with 95% bootstrap confidence intervals. Interpretation follows standard benchmarks: <0.20 slight, 0.21–0.40 fair, 0.41–0.60 moderate, 0.61–0.80 substantial, >0.80 almost perfect. N refers to the cohort in which the comparison was performed: OW comparisons use the full open-weight cohort (N = 4,886); CW comparisons use the closed-weight benchmark cohort (N = 200). The GT inter-agreement section reports pairwise agreement between the three algorithmic ground-truth frameworks and is provided as reference context; notably, DiSSCo GT falls below the moderate threshold against both other GTs, indicating that definitional divergence between severity frameworks is not unique to MOSAIC. The internal pipeline agreement section reports agreement between individual assessor agents (A1, A2) before consolidation, and between each assessor and the final consolidated output; for the CW pipeline, the range reflects two nearly identical values; for the OW pipeline, the range reflects A1 (Gemma, κw = 0.911) and A2 (Qwen, κw = 0.959).*

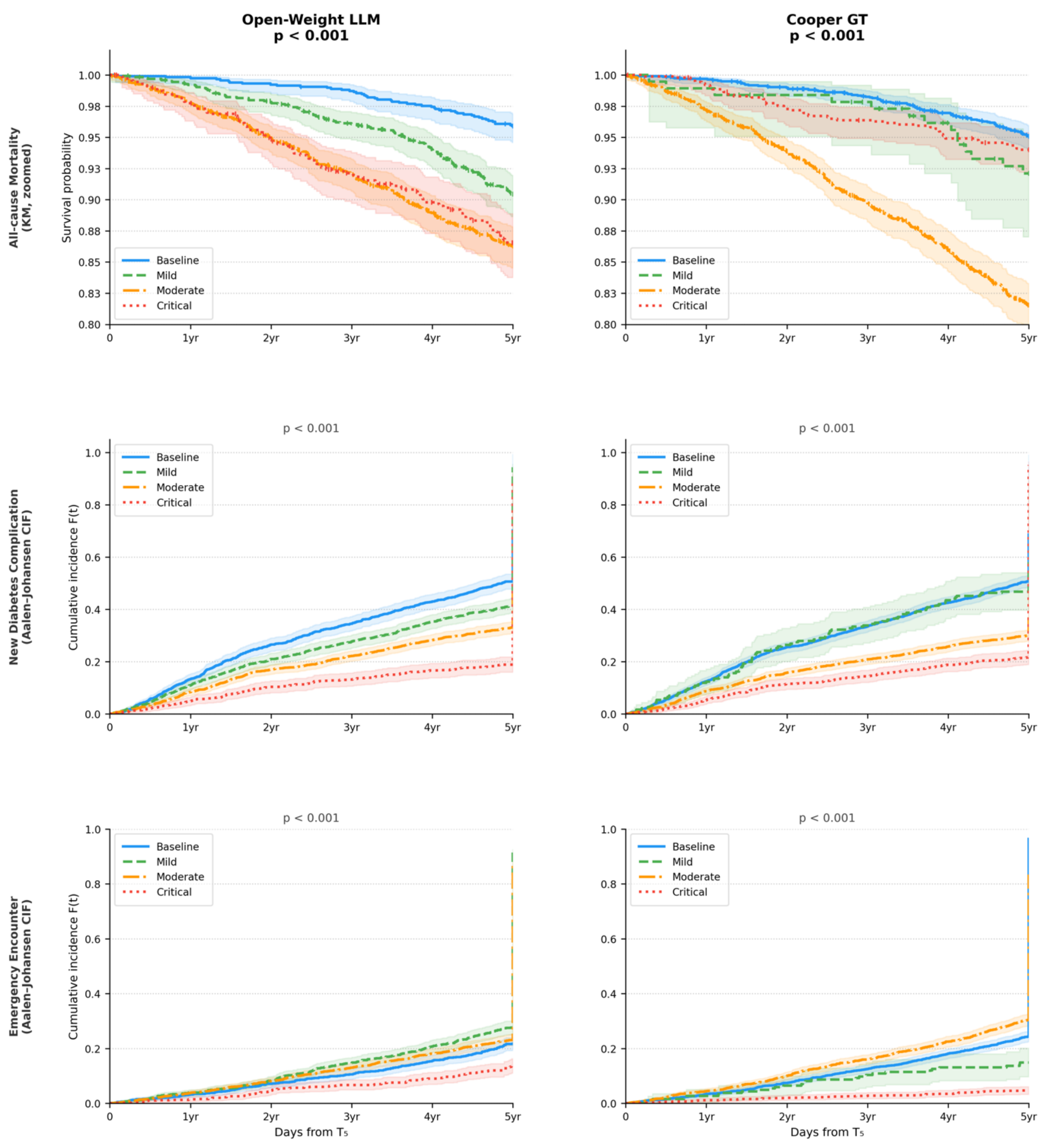

**Figure 2.** Survival and cumulative incidence outcomes by severity tier — Open-Weight MOSAIC and Cooper GT (OW cohort, N = 4,886).

***Legend***: *Each column presents results for one severity classification framework (left: Open-Weight MOSAIC; right: Cooper GT) across three outcomes. Top row: Kaplan–Meier survival curves for all-cause mortality; the y-axis is zoomed to the event range (0.80–1.00) to resolve tier separation. Middle row: cumulative incidence of new diabetes complications estimated using the Aalen–Johansen estimator, with death treated as a competing risk. Bottom row: cumulative incidence of emergency encounters (Aalen–Johansen). In all panels, the x-axis represents days from the* $T_5$ *landmark and shaded regions indicate 95% confidence intervals. Log-rank p-values are shown above each panel. Cooper GT was selected as the comparator because it is the most recently validated ground-truth framework; equivalent curves for Young GT, DiSSCo GT, and MOSAIC Frozen are provided in Supplementary Figure 6.*

**Table 3.** Crude Cox proportional-hazards ratios and restricted mean survival time (RMST) for all-cause mortality by severity tier.

| Tier | N | Events | RMST (days) | Crude HR (95% CI) | p-value |
|---|---|---|---|---|---|
| **Closed-weight benchmark cohort (N = 200)** | | | | | |
| **Young GT (DCSI)** | | | | | |
| Mild | 50 | 5 | 1739.2 | 3.58 (0.93–13.73) | 0.063 |
| Moderate | 50 | 20 | 1483.1 | 17.19 (5.37–55.06) | <0.001 |
| Critical | 21 | 17 | 1048.2 | 50.68 (15.41–166.69) | <0.001 |
| **Cooper GT** | | | | | |
| Mild | 3 | 0 | 1825.0 | 0.42 (0.00–1745.13) † | 0.838 |
| Moderate | 99 | 25 | 1593.6 | 11.51 (3.34–39.67) | <0.001 |
| Critical | 23 | 17 | 1165.6 | 44.22 (12.31–158.85) | <0.001 |
| **DiSSCo GT** | | | | | |
| Mild | 110 | 24 | 1611.9 | 9.12 (2.56–32.41) | <0.001 |
| Moderate | 22 | 18 | 1168.7 | 48.89 (13.23–180.71) | <0.001 |
| **MOSAIC Frozen** | | | | | |
| Mild | 11 | 1 | 1665.2 | 4.34 (0.03–714.93) † | 0.573 |
| Moderate | 61 | 23 | 1475.2 | 449.80 (44.47–4549.68) † | <0.001 |
| Critical | 11 | 9 | 925.7 | 0.00 (0.00–0.03) † | 0.002 |
| **MOSAIC closed-weight (LLM)** | | | | | |
| Mild | 42 | 4 | 1761.5 | 2.61 (0.64–10.70) | 0.182 |
| Moderate | 78 | 24 | 1518.4 | 10.93 (3.43–34.87) | <0.001 |
| Critical | 23 | 14 | 1261.6 | 28.75 (8.49–97.43) | <0.001 |
| **MOSAIC open-weight (LLM, on CW cohort)** | | | | | |
| Mild | 30 | 2 | 1771.2 | 1.58 (0.29–8.65) | 0.595 |
| Moderate | 102 | 26 | 1587.1 | 7.55 (2.22–25.68) | 0.001 |
| Critical | 26 | 14 | 1330.2 | 18.94 (5.26–68.19) | <0.001 |
| **Open-weight cohort (N = 4,886)** | | | | | |
| **Young GT (DCSI)** | | | | | |
| Mild | 1272 | 222 | 1673.1 | 3.16 (2.58–3.88) | <0.001 |
| Moderate | 1267 | 141 | 1726.7 | 1.89 (1.51–2.36) | <0.001 |
| Critical | 202 | 17 | 1743.4 | 1.41 (0.87–2.28) | 0.164 |
| **Cooper GT** | | | | | |
| Mild | 186 | 14 | 1774.9 | 1.34 (0.79–2.26) | 0.279 |
| Moderate | 1856 | 327 | 1670.2 | 3.38 (2.77–4.12) | <0.001 |
| Critical | 887 | 53 | 1771.1 | 1.06 (0.79–1.43) | 0.694 |
| **DiSSCo GT** | | | | | |
| Mild | 2597 | 337 | 1713.9 | 2.47 (2.02–3.02) | <0.001 |
| Moderate | 263 | 56 | 1642.9 | 4.03 (2.95–5.51) | <0.001 |
| **MOSAIC Frozen** | | | | | |
| Mild | 247 | 36 | 1654.8 | 2.63 (1.86–3.72) | <0.001 |
| Moderate | 1339 | 186 | 1705.1 | 1.97 (1.64–2.37) | <0.001 |
| Critical | 76 | 38 | 1302.9 | 9.79 (6.93–13.81) | <0.001 |
| **MOSAIC open-weight (LLM)** | | | | | |
| Mild | 1247 | 117 | 1760.9 | 1.63 (1.26–2.11) | <0.001 |
| Moderate | 1760 | 231 | 1704.0 | 2.43 (1.93–3.07) | <0.001 |
| Critical | 707 | 93 | 1708.8 | 2.34 (1.77–3.09) | <0.001 |

***Legend:*** *Hazard ratios are from unadjusted (crude) Cox models with Baseline severity as the reference category (HR = 1.00). RMST is the restricted mean survival time over a 5-year (1,825-day) horizon, estimated from the tier-specific Kaplan–Meier curve. N = patients in tier; Events = deaths within the tier. p-values < 0.001 are shown as "<0.001".*

*† Estimate unstable: very small stratum and/or sparse events (and, for MOSAIC Frozen Critical, complete separation) produce extreme point estimates and very wide confidence intervals in the closed-weight cohort; interpret with caution and prefer the adjusted models.*

## SUPPLEMENTARY MATERIAL

**Supplementary Table 1.** Clinical Codes Utilized in the Study.

| Code | Vocabulary | Description | Used in |
|---|---|---|---|
| **368009** | SNOMED-CT | Heart valve disorder | DISSCO GT |
| **4855003** | SNOMED-CT | Retinopathy / visual loss | DISSCO GT |
| **6456007** | SNOMED-CT | Atrial fibrillation / flutter | DISSCO GT |
| **13644009** | SNOMED-CT | Hypercholesterolaemia | DISSCO GT, MOSAIC |
| **22298006** | SNOMED-CT | Myocardial infarction | DCSI GT, Cooper GT, DISSCO GT, Complications |
| **34436003** | SNOMED-CT | Albuminuria | DISSCO GT |
| **37130007** | SNOMED-CT | Diabetic foot ulcer / foot complication | Cooper GT, MOSAIC |
| **38341003** | SNOMED-CT | Hypertensive disorder | DISSCO GT |
| **40275004** | SNOMED-CT | Peripheral vascular disease | DISSCO GT, Complications |
| **42343007** | SNOMED-CT | Congestive heart failure | DISSCO GT |
| **44054006** | SNOMED-CT | Type 2 diabetes mellitus (generic) | T2D diagnosis |
| **44808001** | SNOMED-CT | Peripheral vascular disease (variant) | MOSAIC |
| **46177005** | SNOMED-CT | End-stage renal disease | Cooper GT, DISSCO GT, MOSAIC |
| **47801003** | SNOMED-CT | Foot deformity / ulcer | DISSCO GT |
| **49436004** | SNOMED-CT | Atrial fibrillation | DISSCO GT |
| **53741008** | SNOMED-CT | Coronary arteriosclerosis / coronary heart disease | DISSCO GT, MOSAIC |
| **55822004** | SNOMED-CT | Hyperlipidaemia | DISSCO GT, MOSAIC |
| **56819008** | SNOMED-CT | Endocarditis | DISSCO GT |
| **57054005** | SNOMED-CT | Acute myocardial infarction | DISSCO GT |
| **57177007** | SNOMED-CT | Foot ulcer (Zghebi foot-ulcer domain) | DISSCO GT |
| **57772001** | SNOMED-CT | Cardiomyopathy | DISSCO GT |
| **57822003** | SNOMED-CT | Stable angina | DISSCO GT |
| **59621000** | SNOMED-CT | Essential hypertension | DISSCO GT, MOSAIC |
| **79619009** | SNOMED-CT | Heart valve disorder | DISSCO GT |
| **84114007** | SNOMED-CT | Heart failure | Cooper GT, DISSCO GT, MOSAIC |
| **85898001** | SNOMED-CT | Cardiomyopathy | DISSCO GT |
| **90688005** | SNOMED-CT | Chronic renal failure / nephropathy | DISSCO GT |
| **127013003** | SNOMED-CT | Diabetic renal disease / diabetic nephropathy | T2D diagnosis, DCSI GT, Cooper GT, DISSCO GT, MOSAIC, Complications |
| **161695004** | SNOMED-CT | Foot ulcer (Zghebi foot-ulcer domain) | DISSCO GT |
| **175901007** | SNOMED-CT | End-stage renal disease / renal replacement | DISSCO GT |
| **193349004** | SNOMED-CT | Blindness / visual loss | DISSCO GT |
| **193350004** | SNOMED-CT | Diabetic retinopathy (Zghebi retinopathy domain) | DISSCO GT |
| **193387007** | SNOMED-CT | Diabetic retinopathy (Zghebi retinopathy domain) | DISSCO GT |
| **194828000** | SNOMED-CT | Angina pectoris | DCSI GT, Cooper GT, DISSCO GT, MOSAIC, Complications |
| **199049005** | SNOMED-CT | Diabetic neuropathy | Cooper GT, MOSAIC |
| **230572002** | SNOMED-CT | Neuropathy (Zghebi neuropathy domain) | DISSCO GT |
| **230690007** | SNOMED-CT | Stroke / cerebrovascular accident | DCSI GT, Cooper GT, DISSCO GT, MOSAIC |
| **230716006** | SNOMED-CT | Transient ischaemic attack (TIA) | DISSCO GT |
| **232019003** | SNOMED-CT | Diabetic retinopathy (Zghebi retinopathy domain) | DISSCO GT |
| **233829000** | SNOMED-CT | Myocardial infarction / coronary event | DCSI GT, Cooper GT, DISSCO GT, Complications |
| **237636009** | SNOMED-CT | Hypoglycaemia | DCSI GT, DISSCO GT, Complications |
| **266257000** | SNOMED-CT | Transient ischaemic attack (TIA) | DCSI GT, Cooper GT, DISSCO GT, MOSAIC, Complications |
| **271651004** | SNOMED-CT | Amputation | DISSCO GT |
| **299960009** | SNOMED-CT | Neuropathy (Zghebi neuropathy domain) | DISSCO GT |
| **302866003** | SNOMED-CT | Hypoglycaemia | DISSCO GT |
| **372070002** | SNOMED-CT | Gangrene | DISSCO GT |
| **394659003** | SNOMED-CT | Acute coronary syndrome | DISSCO GT |
| **397540003** | SNOMED-CT | Blindness / severe visual loss | DISSCO GT |
| **399957001** | SNOMED-CT | Peripheral vascular disease | DISSCO GT |
| **410429000** | SNOMED-CT | Cardiac arrest | DISSCO GT |
| **413758000** | SNOMED-CT | Cerebrovascular accident / stroke | DISSCO GT, Complications |
| **413844008** | SNOMED-CT | Chronic ischaemic heart disease | MOSAIC |
| **414545008** | SNOMED-CT | Ischaemic / coronary heart disease | DISSCO GT |
| **420422005** | SNOMED-CT | Diabetic ketoacidosis | DCSI GT, DISSCO GT, Complications |
| **421725003** | SNOMED-CT | Hypoglycaemia | DISSCO GT |

| **422034002** | SNOMED-CT | Diabetic retinopathy due to type 2 diabetes mellitus | T2D diagnosis, DCSI GT, Cooper GT, DISSCO GT, MOSAIC, Complications |
|---|---|---|---|
| **422504002** | SNOMED-CT | Ischaemic stroke | DISSCO GT |
| **428575007** | SNOMED-CT | End-stage renal disease | DISSCO GT |
| **431856006** | SNOMED-CT | Diabetic nephropathy | DISSCO GT |
| **698754002** | SNOMED-CT | Peripheral vascular disease | DCSI GT, Cooper GT, DISSCO GT, MOSAIC |
| **707534000** | SNOMED-CT | Dyslipidaemia | MOSAIC |
| **709044004** | SNOMED-CT | Chronic kidney disease | Cooper GT, MOSAIC |
| **719216001** | SNOMED-CT | Hyperosmolar hyperglycaemic state (HHS) | DISSCO GT |
| **981000124106** | SNOMED-CT | Heart failure (Synthea variant code) | MOSAIC, Complications |
| **1501000119109** | SNOMED-CT | Proliferative diabetic retinopathy due to type 2 diabetes mellitus | T2D diagnosis, DCSI GT, Cooper GT, MOSAIC, Complications |
| **1551000119108** | SNOMED-CT | Nonproliferative diabetic retinopathy due to type 2 diabetes mellitus | T2D diagnosis, DCSI GT, Cooper GT, DISSCO GT, MOSAIC, Complications |
| **57171000119105** | SNOMED-CT | Amputation | DISSCO GT |
| **60951000119105** | SNOMED-CT | Blindness due to type 2 diabetes mellitus | T2D diagnosis, DCSI GT, Cooper GT, DISSCO GT, MOSAIC, Complications |
| **90781000119102** | SNOMED-CT | Microalbuminuria due to type 2 diabetes mellitus | T2D diagnosis, DCSI GT, Cooper GT, DISSCO GT, MOSAIC, Complications |
| **97331000119101** | SNOMED-CT | Macular edema due to type 2 diabetes mellitus | T2D diagnosis, DCSI GT, Cooper GT, MOSAIC, Complications |
| **129331000119104** | SNOMED-CT | Amputation (diabetic foot) | Cooper GT, MOSAIC |
| **140101000119109** | SNOMED-CT | Neuropathy | DISSCO GT |
| **157141000119108** | SNOMED-CT | Proteinuria due to type 2 diabetes mellitus | T2D diagnosis, DCSI GT, Cooper GT, DISSCO GT, MOSAIC, Complications |
| **368581000119106** | SNOMED-CT | Neuropathy due to type 2 diabetes mellitus | T2D diagnosis, DCSI GT, Cooper GT, DISSCO GT, MOSAIC |
| **11101003** | SNOMED-CT (procedure) | Coronary intervention (procedure) | DISSCO GT |
| **36969009** | SNOMED-CT (procedure) | Coronary intervention / PCI (procedure) | DISSCO GT |
| **180030006** | SNOMED-CT (procedure) | Amputation procedure | DISSCO GT |
| **406149000** | SNOMED-CT (procedure) | Laser therapy of eye (procedure) | DISSCO GT |
| **415070008** | SNOMED-CT (procedure) | Retinal laser photocoagulation therapy (procedure) | DISSCO GT |
| **50849002** | SNOMED-CT | Emergency room admission | Outcomes |
| **185349003** | SNOMED-CT | Outpatient Encounter | Outcomes |
| **162673000** | SNOMED-CT | Encounter for 'check-up' | Outcomes |
| **270427003** | SNOMED-CT | Encounter for symptom / Encounter for problem | Outcomes |
| **305351004** | SNOMED-CT | Admission to surgical department | Outcomes |
| **71388002** | SNOMED-CT | Patient encounter procedure | Outcomes |
| **11429006** | SNOMED-CT | Consultation for treatment | Outcomes |
| **390906007** | SNOMED-CT | Asthma follow-up | Outcomes |
| **308646001** | SNOMED-CT | Death Certification | Outcomes |
| **32485007** | SNOMED-CT | Non-urgent orthopedic admission (Generic Hospital Admission) | Outcomes |
| **305411003** | SNOMED-CT | Obstetric emergency hospital admission | Outcomes |
| **183452005** | SNOMED-CT | Emergency hospital admission for asthma | Outcomes |
| **106892** | RxNorm | Insulin (treatment-ID anchor) | Treatment ID, Drug class, Cooper GT, MOSAIC |
| **197805** | RxNorm | Sulfonylurea (oral antidiabetic) | Drug class, Cooper GT, MOSAIC |
| **316049** | RxNorm | Sulfonylurea (oral antidiabetic) | Drug class, Cooper GT, MOSAIC |
| **847207** | RxNorm | Insulin (product variant) | Drug class, Cooper GT, MOSAIC |
| **847209** | RxNorm | Insulin (product variant) | Drug class, Cooper GT, MOSAIC |
| **847230** | RxNorm | Insulin (product variant) | Drug class, Cooper GT, MOSAIC |
| **847257** | RxNorm | Insulin (product variant) | Drug class, Cooper GT, MOSAIC |
| **860975** | RxNorm | Metformin (oral antidiabetic; treatment-ID anchor) | Treatment ID, Drug class, Cooper GT, MOSAIC |
| **860978** | RxNorm | Metformin (product/strength variant) | Drug class, Cooper GT, MOSAIC |
| **860981** | RxNorm | Metformin (product/strength variant) | Drug class, Cooper GT, MOSAIC |
| **860996** | RxNorm | Metformin (product/strength variant) | Drug class, Cooper GT, MOSAIC |
| **861004** | RxNorm | Metformin (product/strength variant) | Drug class, Cooper GT, MOSAIC |
| **861006** | RxNorm | Metformin (product/strength variant) | Drug class, Cooper GT, MOSAIC |
| **861010** | RxNorm | Metformin (product/strength variant) | Drug class, Cooper GT, MOSAIC |
| **897122** | RxNorm | Liraglutide (GLP-1 receptor agonist) | Treatment ID |
| **1373463** | RxNorm | Canagliflozin (SGLT2 inhibitor) | Treatment ID |
| **2160-0** | LOINC | Serum creatinine | DCSI GT |
| **14959-1** | LOINC | Urine albumin-to-creatinine ratio (UACR) | DCSI GT, Cooper GT, DISSCO GT, MOSAIC |
| **4548-4** | LOINC | Haemoglobin A1c (HbA1c) | DCSI GT, Cooper GT, MOSAIC |
| **33914-3** | LOINC | Estimated GFR (CKD-EPI) | Cooper GT, MOSAIC |
| **1558-6** | LOINC | Fasting plasma glucose | MOSAIC |
| **1521-4** | LOINC | Postprandial glucose (2h) | MOSAIC |
| **39156-5** | LOINC | Body mass index | MOSAIC |
| **20448-7** | LOINC | Fasting insulin | MOSAIC |
| **1986-9** | LOINC | C-peptide | MOSAIC |
| **18262-6** | LOINC | LDL cholesterol | MOSAIC |
| **2093-3** | LOINC | Total cholesterol | MOSAIC |
| **2571-8** | LOINC | Triglycerides | MOSAIC |
| **2085-9** | LOINC | HDL cholesterol | MOSAIC |

*Concept codes used across the MOSAIC pipeline and the three reference frameworks. The "Used in" column lists every component each code feeds. Codes recur across uses (e.g. one SNOMED concept may serve diagnosis, a ground truth, and MOSAIC). Diagnosis cohort codes (SNOMED) and treatment-ID anchor drugs (Metformin 860975, Canagliflozin, Liraglutide, Insulin 106892) define Table 3 cohort selection. Drug-class codes (DRUG_CLASSES) feed the treatment-group definition and are also consumed by Cooper (medication count / insulin) and MOSAIC (oral-antidiabetic count + insulin). Vocabularies: SNOMED-CT (diagnoses/procedures), RxNorm (medications), LOINC (laboratory tests). GT = ground truth; DCSI = Diabetes Complications Severity Index (Young et al.); DISSCO = Diabetes Severity Score (Zghebi et al.); MOSAIC = frozen rule-based MOSAIC classifier; T2D(M) = type 2 diabetes (mellitus); OAD = oral antidiabetic; HbA1c = glycated haemoglobin; eGFR = estimated glomerular filtration rate; UACR = urine albumin-to-creatinine ratio; PPG = postprandial glucose; BMI = body mass index; LDL/HDL = low-/high-density lipoprotein; SGLT2i = sodium–glucose cotransporter-2 inhibitor; GLP-1 = glucagon-like peptide-1 receptor agonist; PCI = percutaneous coronary intervention; TIA = transient ischaemic attack; HHS = hyperosmolar hyperglycaemic state.*

**Supplementary Table 2**. Phase 1 Closed-Weight Agent Pipeline Configuration.

| Agent Role | Model | Task Prompt | Expected Output | Context Input |
|---|---|---|---|---|
| **Assessor 1 & 2** <br> ***(Senior Clinical Researcher)*** | deepseek-chat <br> gpt-4o | "Search for T2D severity phenotyping criteria considering comorbidities and natural history. Return a bulleted list of the top most important criteria for which there is medical consensus. Make them operable with specific thresholds." | "A brief summary of clinical criteria for T2D severity along with references, optimised for EHR." | **Tavily Query:** "T2D severity criteria comorbidities natural history clinical guidelines" |
| **Consolidator** <br> ***(Chief Medical Informatician)*** | claude-sonnet-4-6 | "Review the research provided by the two clinical researchers. Compare their findings, identify the strongest consensus, and produce a final, unified framework for T2D severity phenotyping." | "A definitive, consolidated guide of operable EHR criteria for T2D severity with specific thresholds and references." | **Parallel Context:** Receives the combined outputs from both Assessor 1 and Assessor 2. |

**Supplementary Table 3.** LLM identifiers, roles, weight type, and inference settings across both MOSAIC phases.

| Model (Role) | Developer / Version | Weight Type | Approx. VRAM (4-bit) | Temperature |
|---|---|---|---|---|
| DeepSeek-Chat (Assessor 1) | DeepSeek; deepseek-chat (V2) | Closed-weight (API) | N/A | 0.0 |
| GPT-4o (Assessor 2) | OpenAI; gpt-4o (2024-08-06) | Closed-weight (API) | N/A | 0.0 |
| claude-sonnet-4-6 (Consolidator) | Anthropic; claude-sonnet-4-6 | Closed-weight (API) | N/A | 0.0 |
| Llama 3.1 70B (Consolidator) | Meta; llama3.1:70b (Ollama Q4) | Open-weight (local) | ~35 GB | 0.0 |
| Gemma 2 27B (Assessor A) | Google DeepMind; gemma2:27b (Ollama Q4) | Open-weight (local) | ~14 GB | 0.0 |
| Qwen 2.5 14B (Assessor B) | Alibaba; qwen2.5:14b (Ollama Q4) | Open-weight (local) | ~8 GB | 0.0 |

***Legend:*** *This table outlines the specific language models utilized across both phases of the MOSAIC framework, detailing their assigned roles, weight accessibility (closed API vs. open local), memory requirements for local execution, and deterministic generation settings. Abbreviations: API = Application Programming Interface; GB = Gigabytes; LLM = Large Language Model; N/A = Not Applicable (local memory is not required for API models); Q4 = 4-bit Quantization (a technique to reduce model size); VRAM = Video Random Access Memory.*

**Supplementary Table 4.** Ollama server environment variables used for the open-weight framework [CHART 4b; TRIPOD-LLM 8].

| Environment Variable | Value / Purpose |
|---|---|
| OLLAMA_NUM_GPU | 99 — allocate all available GPU layers |
| OLLAMA_FLASH_ATTENTION | 1 — enable Flash Attention 2 for throughput |
| OLLAMA_KV_CACHE_TYPE | q8_0 — quantised key-value cache to reduce VRAM pressure |
| OLLAMA_MAX_LOADED_MODELS | 3 — all three models resident in VRAM simultaneously |
| OLLAMA_KEEP_ALIVE | -1 — models never unloaded between calls |
| OLLAMA_NUM_PARALLEL | 1 — single-slot sequential inference |

**Supplementary Table 5.** Phase 2 OW Agent Pipeline Configuration.

| Component | Architecture and Prompt Templates |
|---|---|
| **Dr. A — Clinical Informatician** | **System Persona (Gemma):** Clinical informatician specialising in EHR-based T2D phenotyping. Conservative—only count what is explicitly documented.<br>**Expected Output:** Brief domain assessment, final tier, binary status, confidence, and four scores. |
| **Dr. B — Consultant Endocrinologist** | **System Persona (Qwen):** Consultant endocrinologist with 20 years T2D experience. Considers the full 5-year clinical trajectory and takes a holistic, outcomes-focused approach.<br>**Expected Output:** Brief domain assessment, final tier, binary status, confidence, and four scores. |
| **Assessor Task Prompt** | **Shared User Prompt (Run Concurrently):**<br>Classify this T2D patient's severity at their $T_5$ (5-year) reclassification point.<br>*TIERS:* Tier 1=Baseline T2D \| Tier 2=Mild Complications \| Tier 3=Moderate Complications \| Tier 4=Advanced/Critical<br><br>*Rule:* Classify based on ALL available evidence—medications, procedures, conditions, and lab values—not just explicit complication codes. Only default to Baseline T2D if the record is truly empty of any complication signals. Never respond with "Insufficient Data".<br><br>=== FRAMEWORK ===<br>{CONCONSOLIDATED_FRAMEWORK}<br><br>=== PATIENT RECORD ===<br><br><br>{raw_record}<br>*Response Format:*<br>1. Domain assessment (1-2 sentences per domain only). Explicitly state: "Concurrent Tier 3 domains identified: [list them]" and "Rule 0a check: [met / not met]".<br>2. Notable clinical observations not covered by framework.<br>3. Final tier + binary classification (COMPLEX or NOT_COMPLEX) + confidence (High/Medium/Low).<br>4. Scores (0-100 each, summing to ~100) for all 4 tiers. |
| **Chief Medical Informatician** | **System Persona (Llama):** Final Arbiter responsible for reviewing both assessors' classifications, resolving disagreements, and generating the final clinical severity phenotype.<br>**Expected Output:** ===FINAL=== block followed by 1-2 sentences of resolution reasoning. |
| **Consolidator Task Prompt** | **Sequential Arbitration User Prompt:**<br>Review the two assessors' outputs for patient {patient_id} and produce the final classification. Before finalising, perform a structured self-check grounded in the framework's own decision logic:<br>1. *Evidence Synthesis:* Count distinct domain criteria at Tier 3 or above across both files combined. |
| | 2. *Rule 0a Check:* If $\ge 2$ concurrent Tier 3 domains are present, mandate upward reclassification to Tier 4.<br>3. *Rule 0 Check:* Enforce highest single triggered tier—not an average—unless explicitly contradicted.<br>4. *Under-classification Check:* Confirm insulin dependence + complication is correctly weighted via Domain 7 proxy.<br>5. *Over-classification Check:* Confirm Tier 4 assignment is not based on a single isolated finding.<br>Output the ===FINAL=== block first, then write 2-3 sentences summarising your self-check.<br>===FINAL===<br>PATIENT_ID: {patient_id} \| FOUR_TIER: [...] \| BINARY: [...] \| ASSESSOR_A_TIER: [...] \| ASSESSOR_B_TIER: [...] \| CONFIDENCE: [...] \| AGREEMENT: [...] \| INDEX_DATE_CONTEXT: {INDEX_DATE_LABEL} \| KEY_EVIDENCE: [...] \| SCORE_BASELINE_T2D to SCORE_ADVANCED_CRITICAL (0-100 each)<br>===END=== |

**Supplementary Table 6.** Phase 2 CW Agent Pipeline Configuration.

| Component | Architecture and Prompt Templates |
|---|---|
| **Dr. A Clinical Informatician** | System Persona (GPT-4o): Methodical and conservative clinical informatician specializing in EHR-based phenotyping. Works systematically through framework domains but evaluates broader clinical trajectories (encounter frequency, care plan activity, medication escalation). Never refuses to classify sparse records; treats absence of complications as an explicit finding indicating Baseline T2D.<br>Expected Output: Structured assessment containing domain evidence, beyond-framework insights, classifications, confidence, and independent tier support scores. |
| **Dr. B Consultant Endocrinologist** | System Persona (DeepSeek): Consultant endocrinologist with 20 years of clinical experience. Takes a holistic, outcomes-focused approach across the full 5-year window. Looks "between the lines" (e.g., matching care burden and escalation patterns against lab values). Emphasizes that an absence of complications is a definitive clinical finding.<br>Expected Output: Identical to Dr. A (Structured domain breakdown, holistic clinical observations, four-tier and binary statuses, confidence, and four tier support scores). |
| **Assessor Task Prompt** | Shared User Prompt (Executed Concurrently & Asynchronously):<br><br>Classify a Type 2 Diabetes patient's severity at their $T_5$ reclassification point (5 years after first T2D treatment) using the full covariate window data.<br>*Tiers:* Tier 1: Baseline T2D \| Tier 2: Mild Complications \| Tier 3: Moderate Complications \| Tier 4: Advanced/Critical<br><br>*Sparse Record Directive:* All patients have confirmed T2D. If records are sparse, do not select "Insufficient Data." Absence of complications is a valid finding; classify as "Baseline T2D."<br><br>=== CONSOLIDATED SEVERITY PHENOTYPING FRAMEWORK ===<br>{CONSOLIDATED_FRAMEWORK}<br>=== PATIENT CLINICAL RECORD ($T_5$ covariate window) ===<br>{raw_record}<br><br>*Instructions:*<br>1. *Framework Domains:* Systematically detail evidence, thresholds, and triggered tiers for each domain.<br>2. *Beyond the Framework:* Evaluate medication trajectories (e.g., insulin initialization), encounter types (ED/hospitalizations), comorbidity clustering, active care plans, and lab trends (e.g., worsening eGFR).<br>3. *Rule Enforcement:* Apply Rule 0 (highest single domain tier) and Rule 0a (\$ (\$\g\$ concurrent Tier 3 domains \$s \$\rightar\$ Tier 4).<br>*Required Output Format:* Domain assessment \$t \$\rightar\$ Beyond-framework observations \$s \$\rightar\$ Met vs unmet criteria summary \$y \$\rightar\$ Four-Tier status \$s \$\rightar\$ Binary Status (COMPLEX vs |
| | NOT_COMPLEX) $) $\rightar$ Confidence $e $\rightar$ Gaps $s $\rightar$ Independent Tier Support Scores (0 to 100 each). |
| Chief Medical Informatician | System Persona (Claude 3.5 Sonnet): Final Arbiter responsible for pipeline data reconciliation. Identifies inter-rater agreement/disagreement, parses specific evidence citations when assessors conflict, and determines the best-supported clinical interpretation. Re-enforces baseline rules for sparse data profiles.<br>Expected Output: Unified ===FINAL=== block printed at the absolute beginning of the stream, followed by qualitative resolution reasoning. |
| Consolidator Task Prompt | Sequential Arbitration User Prompt:<br>Review the parallel evaluations for patient {patient_id} at their $T_5$ point. Sparse records default to Baseline T2D.<br>============================================================<br>CRITICAL INSTRUCTION: OUTPUT THE ===FINAL=== BLOCK FIRST.<br>Write it as the VERY FIRST THING in your response, before any analysis.<br>============================================================<br>===FINAL===<br>PATIENT_ID: {patient_id} \| FOUR_TIER: [...] \| BINARY: [...] \| ASSESSOR_A_TIER/BINARY: [...] \| ASSESSOR_B_TIER/BINARY: [...] \| CONFIDENCE: [...] \| AGREEMENT: Full/Partial/None \| INDEX_DATE_CONTEXT: {INDEX_DATE_LABEL} \| KEY_EVIDENCE: [One decisive sentence] \| SCORE_BASELINE_T2D to SCORE_ADVANCED_CRITICAL (0-100 independent ratings)<br>===END===<br><br>*Post-Block Requirements:* Provide a brief analysis of rater alignment, core evidence drivers, tier score rationales, and Synthea data quality limitations. |

**Supplementary Table 7.** TRIPOD-LLM Checklist.

| Section | Item | Description | Page no. |
|---|---|---|---|
| **Title** | 1 | Identify the study as developing, fine-tuning and/or evaluating the performance of an LLM, specifying the task, the target population and the outcome to be predicted. | i |
| **Introduction** | | | |
| **Abstract** | 2 | See TRIPOD-LLM for abstracts. | iv |
| **Background** | 3a | Explain the healthcare context/use case (for example, administrative, diagnostic, therapeutic and clinical workflow) and rationale for developing or evaluating the LLM, including references to existing approaches and models. | 1-5 |
| | 3b | Describe the target population and the intended use of the LLM in the context of the care pathway, including its intended users in current gold standard practices (for example, healthcare professionals, patients, public or administrators). | 1-5 |
| **Objectives** | 4 | Specify the study objectives, including whether the study describes the initial development, fine-tuning or validation of an LLM (or multiple stages). | 1-2, 6-7 |
| **Methods** | | | |
| **Data** | 5a | Describe the sources of data separately for the training, tuning and/or evaluation datasets and the rationale for using these data (for example, web corpora, clinical research/trial data, EHR data or unknown). | 7 |
| | 5b | Describe the relevant data points and provide a quantitative and qualitative description of their distribution and other relevant descriptors of the dataset (for example, source, languages and countries of origin). | 7 |
| | 5c | Specifically state the date of the oldest and newest item of text used in the development process (training, fine-tuning and reward modeling) and the evaluation datasets. | 7 |
| | 5d | Describe any data preprocessing and quality checking, including whether this was similar across text corpora, institutions and relevant sociodemographic groups. | 11 |
| | 5e | Describe how missing and imbalanced data were handled and provide reasons for omitting any data. | 11 |
| **Analytical Methods** | 6a | Report the LLM name, version and last date of training. | 11, 41-43 |
| | 6b | Report details of the LLM development process, such as LLM architecture, training, fine-tuning procedures and alignment strategy (for example, reinforcement learning and | 11, 41-43 |
| | | direct preference optimization) and alignment goals (for example, helpfulness, honesty and harmlessness). | |
| | 6c | Report details of how the text was generated using the LLM, including any prompt engineering (including consistency of outputs), and inference settings (for example, seed, temperature, max token length and penalties), as relevant. | 11, 41-43 |
| | 6d | Specify the initial and postprocessed output of the LLM (for example, probabilities, classification and unstructured text). | 11, 41-43 |
| | 6e | Provide details and rationale for any classification and, if applicable, how the probabilities were determined and thresholds identified. | 11 |
| **LLM Output** | 7a | Include metrics that capture the quality of generative outputs, such as consistency, relevance, accuracy and presence/type of errors compared to gold standards. | 11 |
| | 7b | Report the outcome metrics' relevance to the downstream task at deployment time and, where applicable, the correlation of metric to human evaluation of the text for the intended use. | 14-17 |
| | 7c | Clearly define the outcome, how the LLM predictions were calculated (for example, formula, code, object and API), the date of inference for closed-source LLMs and evaluation metrics. | 11,14-17 |
| | 7d | If outcome assessment requires subjective interpretation, describe the qualifications of the assessors, any instructions provided, relevant information on demographics of the assessors and inter-assessor agreement. | NA |
| | 7e | Specify how performance was compared to other LLMs, humans and other benchmarks or standards. | 26 |
| **Annotation** | 8a | If annotation was done, report how the text was labeled, including providing specific annotation guidelines with examples. | NA |
| | 8b | If annotation was done, report how many annotators labeled the dataset(s), including the proportion of data in each dataset that was annotated by more than one annotator, and the inter-annotator agreement. | NA |
| | 8c | If annotation was done, provide information on the background and experience of the annotators or the characteristics of any models involved in labeling. | NA |
| **Prompting** | 9a | If research involved prompting LLMs, provide details on the processes used during prompt design, curation and selection. | 37, 41-42 |
| | 9b | If research involved prompting LLMs, report what data were used to develop the prompts. | NA |
| **Summariza-tion** | 10 | Describe any preprocessing of the data before summarization. | |
| **Instruction tuning/ alignment** | 11 | If instruction tuning/alignment strategies were used, what were the instructions, data and interface used for evaluation, and what were the characteristics of the populations doing the evaluation? | NA |
| **Compute** | 12 | Report compute, or proxies thereof (for example, time on what and how many machines, cost on what and how many machines, inference time, floating-point operations per second), required to carry out methods. | 11 |
| **Ethical approval** | 13 | Name the institutional research board or ethics committee that approved the study and describe the participant-informed consent or the ethics committee waiver of informed consent. | NA |
| **Open science** | 14a | Give the source of funding and the role of the funders for the present study. | NA |
| | 14b | Declare any conflicts of interest and financial disclosures for all authors. | NA |
| | 14c | Indicate where the study protocol can be accessed or state that a protocol was not prepared. | NA |
| | 14d | Provide registration information for the study, including register name and registration number, or state that the study was not registered. | NA |
| | 14e | Provide details of the availability of the study data. | 7 |
| | 14f | Provide details of the availability of the code to reproduce the study results. | 14 |
| **Public involvement** | 15 | Provide details of any patient and public involvement during the design, conduct, reporting, interpretation or dissemination of the study or state no involvement. | NA |
| **Results** | | | |
| **Participants** | 16a | When using patient/EHR data, describe the flow of text/EHR/patient data through the study, including the number of documents/questions/participants with and without the outcome/label and follow-up time as applicable. | 7, 9 |
| | 16b | When using patient/EHR data, report the characteristics overall and for each data source or setting and development/evaluation splits, including the key dates, key characteristics and sample size. | 7, 9, 17 |
| | 16c | For LLM evaluation that includes clinical outcomes, show a comparison of the distribution of important clinical variables that may be associated with the outcome between development and evaluation data, if available. | 17 |

| | 16d | When using patient/EHR data, specify the number of participants and outcome events in each analysis (for example, for LLM development, hyperparameter tuning and LLM evaluation). | 17 |
|---|---|---|---|
| **Performance** | 17 | Report LLM performance according to prespecified metrics (see item 7a) and/or human evaluation (see item 7d). | 17-20 |
| **LLM updating** | 18 | If applicable, report the results from any LLM updating, including the updated LLM and subsequent performance. | NA |
| **Discussion** | | | |
| **Interpretation** | 19a | Give an overall interpretation of the main results, including issues of fairness in the context of the objectives and previous studies. | 25-31 |
| **Limitations** | 19b | Discuss any limitations of the study and their effects on any biases, statistical uncertainty and generalizability. | 31 |
| **Usability of the LLM in context** | 19c | Describe any known challenges in using data for the specified task and domain context with reference to representation, missingness, harmonization and bias. | 31 |
| | 19d | Define the intended use for the implementation under evaluation, including the intended input, end-user and level of autonomy/human oversight. | 5 |
| | 19e | If applicable, describe how poor quality or unavailable input data should be assessed and handled when implementing the LLM; that is, what is the usability of the LLM in the context of current clinical care. | 31 |
| | 19f | If applicable, specify whether users will be required to interact in the handling of the input data or use of the LLM, and what level of expertise is required of users. | NA |
| | 19g | Discuss any next steps for future research, with a specific view of the applicability and generalizability of the LLM. | 31 |

*M = LLM methods; D = de novo LLM development; E = LLM evaluation; H = LLM evaluation in healthcare settings; C = classification; OF = outcome forecasting; QA = long-form question answering; IR = information retrieval; DG = document generation; SS = summarization and simplification; MT = machine translation; API = application programming interface.*

**Supplementary Table 8.** CHART Checklist.

| Heading | No | Chart checklist item | Page no. |
|---|---|---|---|
| **Title and Abstract** | | | |
| **Title** | 1a | State that the study is assessing one or more generative AI-driven chatbots for clinical evidence or health advice. | i |
| **Abstract/ Summary** | 1b | Apply a structured format, if applicable. | iv |
| **Introduction** | | | |
| **Background** | 2a | State the scientific background, rationale, and healthcare context for evaluating the generative AI-driven chatbot(s), referencing relevant literature when applicable. | 2-5 |
| | 2b | State the aims and research questions including the target audience, intervention, comparator(s), and outcome(s). | 2 |
| **Methods** | | | |
| **Model Identifiers** | 3a | State the name and version identifier(s) of the generative AI model(s) and chatbot(s) under evaluation, as well as their date of release or last update. | 11, 41-34 |
| | 3b | State whether the generative AI model(s) and chatbot(s) are open-source or closed-source/proprietary. | 11, 41-34 |
| **Model Details** | 4a | State whether the generative AI model was a base model or a novel base model, tuned model, or fine-tuned model. | 11, 41-34 |
| | 4b | If a base model is used, cite its development in sufficient detail to identify the model. | 11, 41-34 |
| | 4c | If a novel base model, tuned model, or fine-tuned model is used, describe the pre- and/or post-implementation/deployment data and parameters. | NA |
| **Prompt Engineering** | 5a | Describe the evolution of study prompt development. | 37, 41-42 |
| | 5ai | Describe the sources of prompts. | 37, 41-42 |
| | 5aii | State the number and characteristics of the individual(s) involved in prompt engineering. | NA |
| | 5aiii | Provide details of any patient and public involvement during prompt engineering. | NA |
| | 5b | Provide study prompts. | NA |
| **Query Strategy** | 6a | State route of access to generative AI model. | 11, 41-34 |
| | 6b | State the date(s) and location(s) of queries for the generative AI-driven chatbot(s) including the day, month, and year as well as city and country. | NA |
| | 6c | Describe whether prompts were input into separate chat session(s). | NA |
| | 6d | Provide all generative AI-driven chatbot output/responses | 14 |
| **Prompting** | 7a | Define the ground truth or reference standard used to define successful generative AI-driven chatbot performance. | 8 |
| | 7b | Describe the process undertaken for generative AI-driven chatbot performance evaluation. | 11 |
| | 7bi | State the number and characteristics of team members involved in performance evaluation. | NA |
| | 7bii | Provide details of any patients and public involvement during the evaluation process. | NA |
| | 7biii | State whether evaluators were blinded to the identity of the generative AI-driven chatbot(s) under assessment. | NA |
| **Sample Size** | 8 | Report how the sample size was determined. | 17 |
| **Data Analysis** | 9a | Describe statistical analysis methods, including any evaluation of reproducibility of generative AI-driven chatbot responses. | 11 |
| | 9ai | Report the measures used for performance evaluation. | 11 |
| **Results** | | | |
| | 10a | Report the performance evaluation undertaken including the alignment between generative AI-driven chatbot output and ground truth or reference standard using quantitative or mixed methods approaches as applicable. | 17-20 |
| | 10b | For responses deviating from the ground truth or reference standard, state the nature of the difference(s). | 27 |
| | 10c | Indicate where the study protocol can be accessed or state that a protocol was not prepared. | 14 |
| **Discussion** | | | |
| | 11a | Interpret study findings in the context of relevant evidence. | 25-31 |
| | 11b | Describe the strengths and limitations of the study. | 31 |
| | 11c | Describe the potential implications for practice, education, policy, regulation, and research. | 1, 31 |
| **Open Science** | | | |
| **Disclosures** | 12a | Report any relevant conflicts of interest for all authors. | NA |
| **Funding** | 12b | Report sources of funding and their role in the conduct and reporting of the study. | NA |
| **Ethics** | 12c | Describe the process undertaken for ethical approval. | 7 |
| | 12ci | Describe the measures taken to safeguard data privacy of patient health information, as applicable. | NA |
| | 12cii | State whether permission/licensing was obtained for the use of original, copyrighted data. | NA |
| **Protocol** | 12d | Provide a study protocol. | 14 |
| **Data availability** | 12e | State where study data, code repository, and model parameters can be accessed. | 14 |

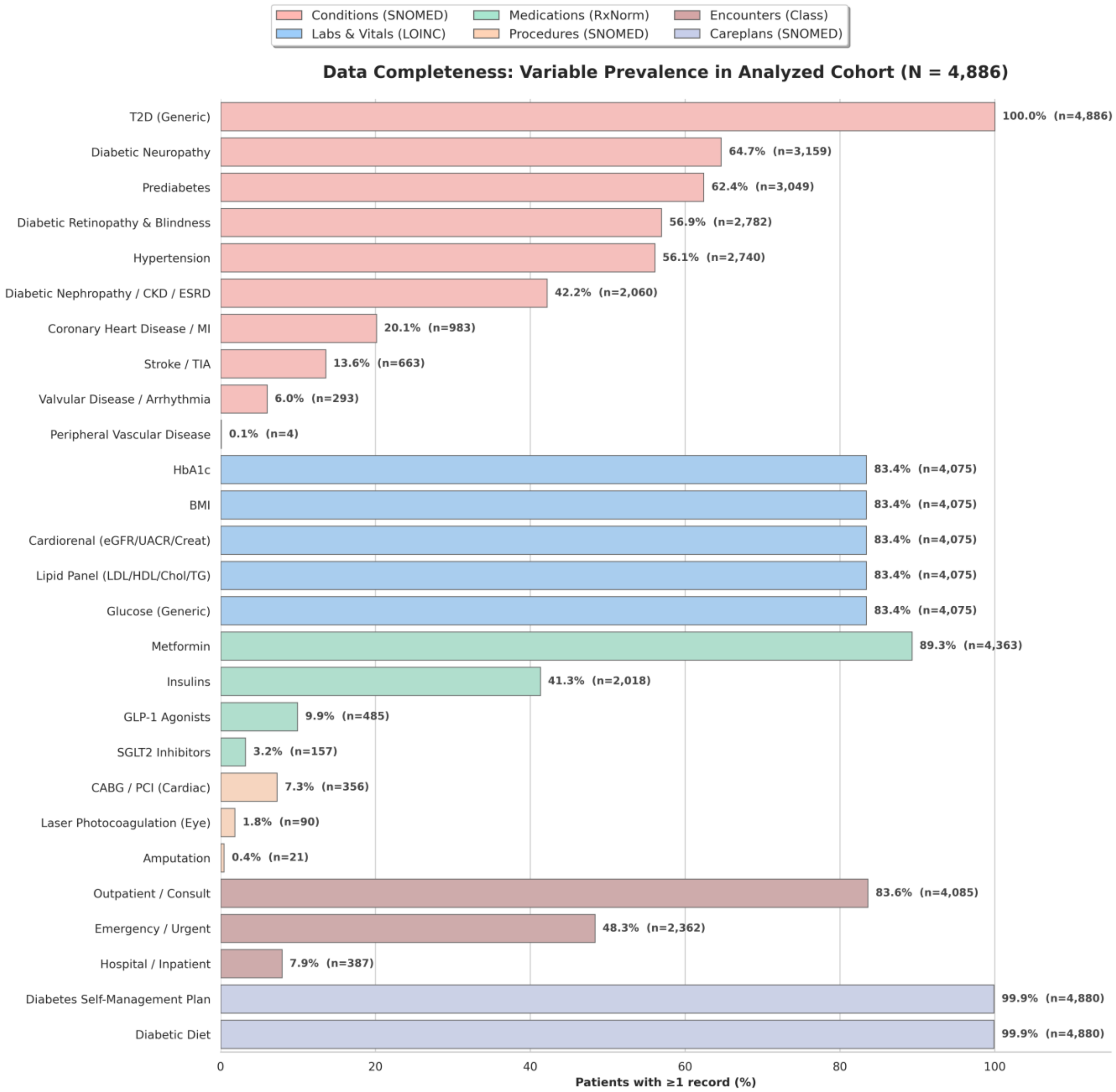

**Supplementary Figure 1**. Vocabulary coverage of clinical concepts used for T2D severity phenotyping in the analysed cohort.

*The figure shows the proportion of patients in the analysed open-weight cohort (N = 4,886) with at least one qualifying record for each clinical concept used in cohort construction, ground-truth derivation, or MOSAIC evidence extraction. Concepts are grouped by data source and vocabulary type, including conditions encoded using SNOMED CT, laboratory and vital-sign measurements encoded using LOINC, medications encoded using RxNorm, procedures encoded using SNOMED CT, encounter classes, and care plans. Bar labels show the percentage of patients with at least one corresponding record and the absolute number of patients. High coverage of core T2D, laboratory, medication, complication, and care-plan signals supports the use of SyntheticMass for structured severity phenotyping. Sparse coverage of peripheral vascular disease, amputation, laser photocoagulation, SGLT2 inhibitors, and some advanced procedure-based markers indicates representational limitations for selected severity domains.*

***Abbreviations:*** *T2D, type 2 diabetes; CKD, chronic kidney disease; ESRD, end-stage renal disease; MI, myocardial infarction; TIA, transient ischemic attack; HbA1c, glycated haemoglobin; BMI, body mass index; eGFR, estimated glomerular filtration rate; UACR, urine albumin-to-creatinine ratio; Creat, creatinine; LDL, low-density lipoprotein; HDL, high-density lipoprotein; Chol, cholesterol; TG, triglycerides; GLP-1, glucagon-like peptide-1; SGLT2, sodium-glucose cotransporter-2; CABG, coronary artery bypass grafting; PCI, percutaneous coronary intervention.*

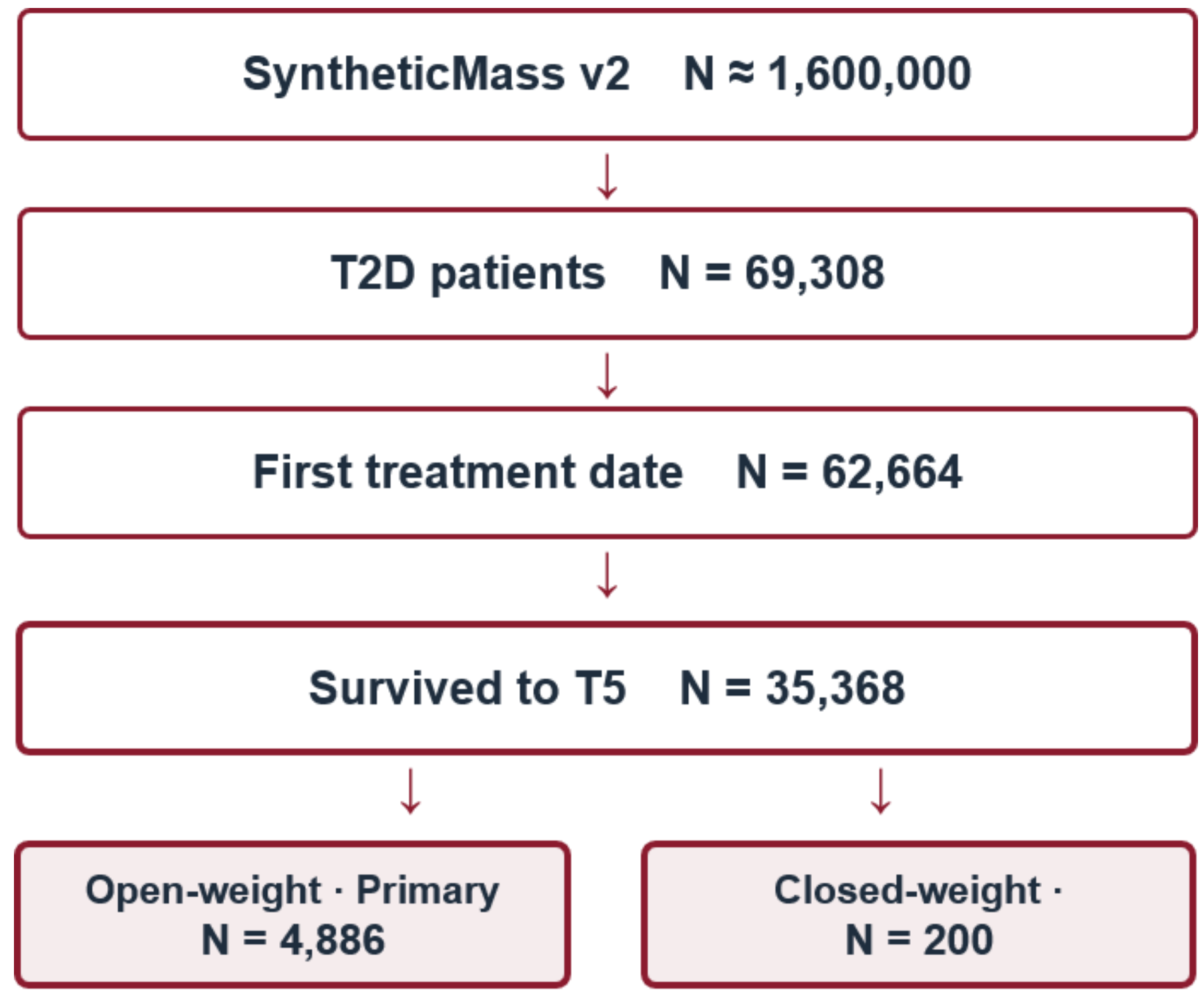

**Supplementary Figure 2**. Study population selection flowchart.

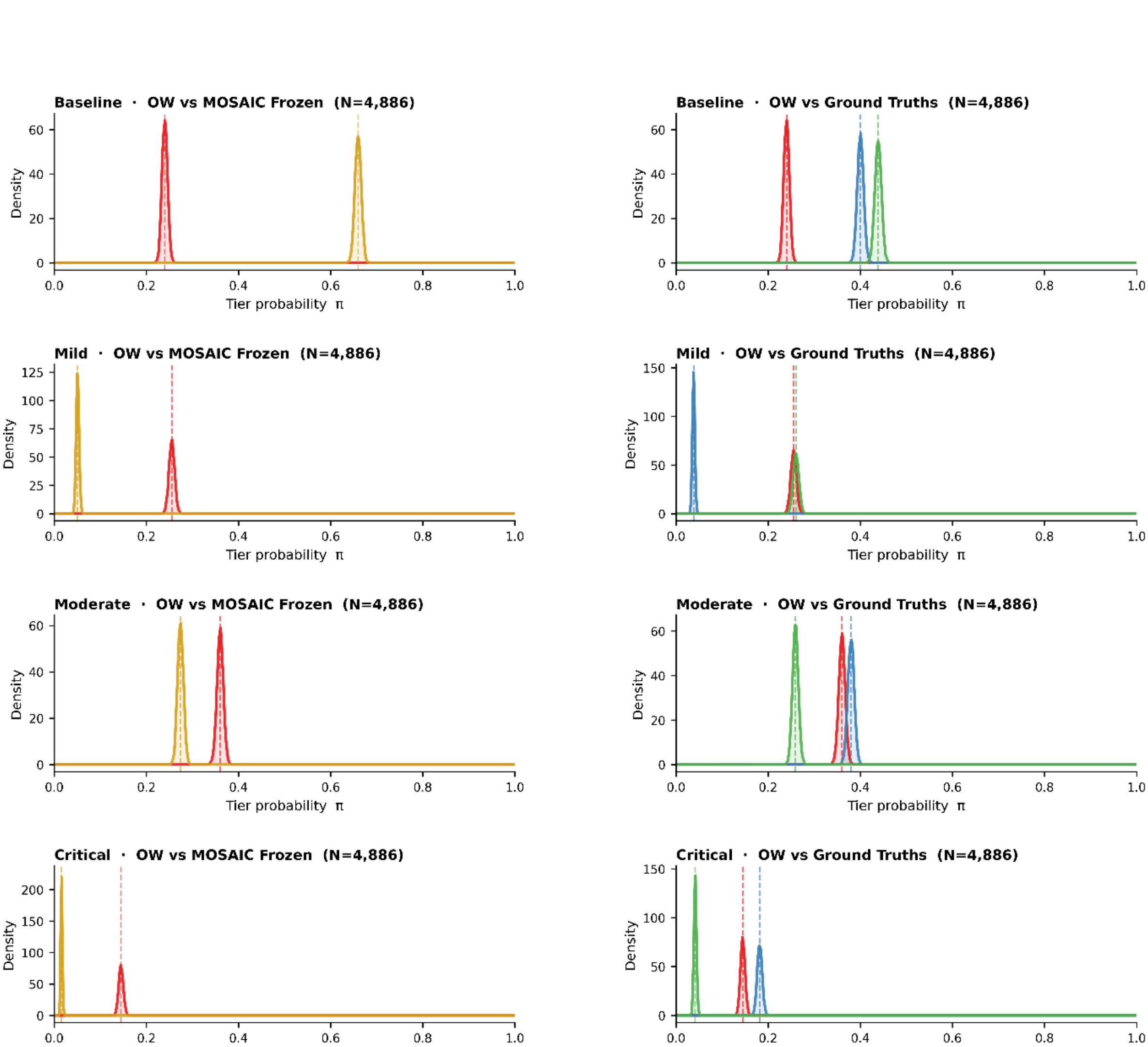

**Supplementary Figure 3.** Posterior tier probability distributions for open-weight MOSAIC compared with MOSAIC Frozen and external ground truths.

***Legend:*** *Posterior tier probabilities were estimated using a Bayesian Dirichlet-Multinomial model in the open-weight cohort. Each row corresponds to one severity tier: Baseline, Mild, Moderate, and Critical. The left column compares open-weight MOSAIC with MOSAIC Frozen. The right column compares open-weight MOSAIC with the external Young GT and Cooper GT frameworks. The x-axis represents the posterior probability, π, that a patient in the cohort is assigned to the corresponding severity tier. The y-axis represents posterior density, showing how concentrated the posterior samples are around each probability value. Solid curves represent posterior kernel density estimates, dashed vertical lines indicate posterior means, and shaded regions represent 95% highest density intervals. Separation between posterior distributions indicates systematic population-level differences in tier assignment, whereas overlapping distributions indicate closer population-level alignment. DiSSCo GT is not displayed in this figure because it assigned zero patients to the Critical tier in this cohort, producing a degenerate posterior distribution concentrated at zero in that panel and distorting the density scale for visual comparison*

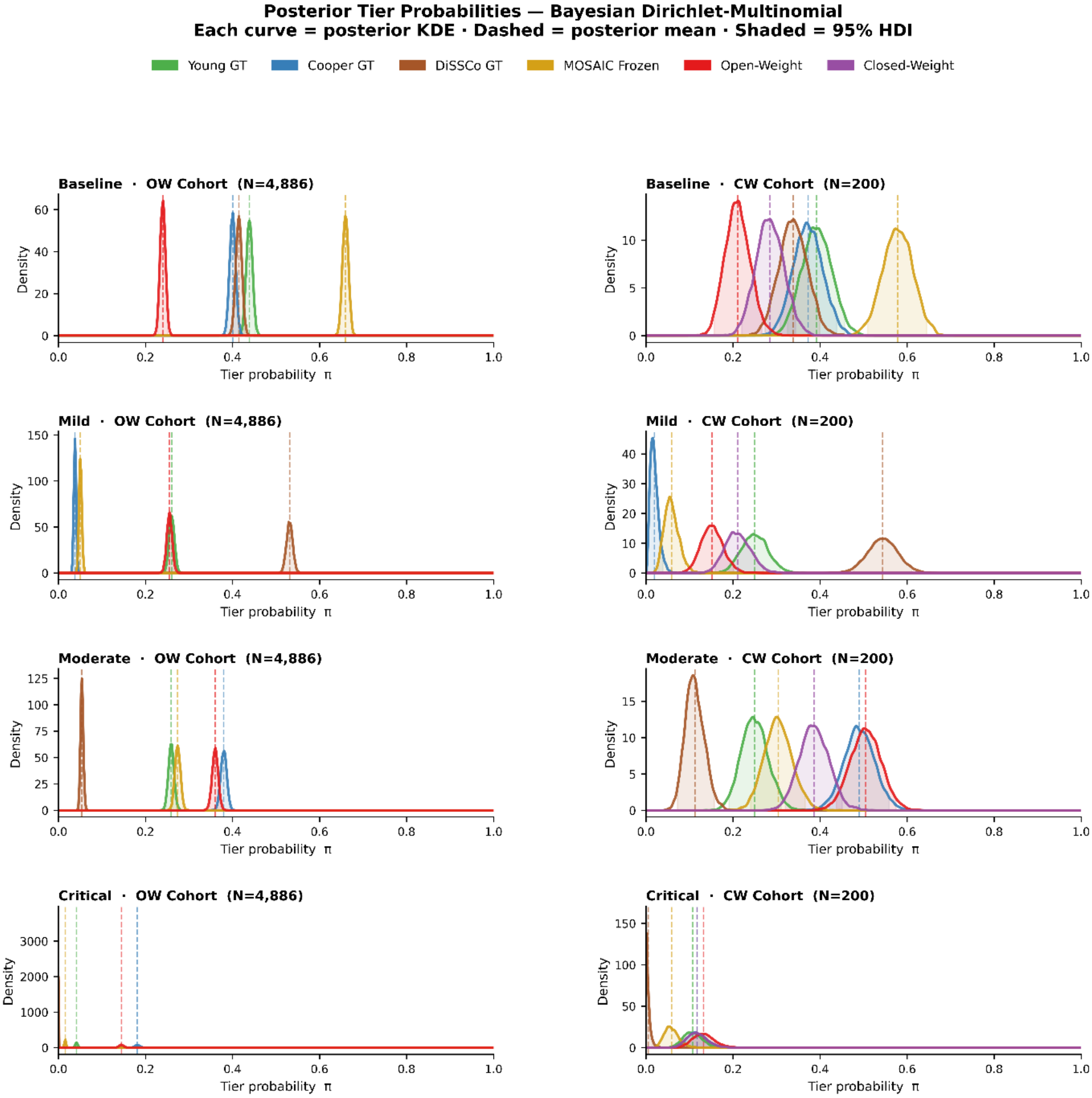

**Supplementary Figure 4**. Posterior tier probability distributions across all classifiers in the open-weight and closed-weight cohorts.

***Legend:*** *Posterior tier probabilities were estimated using a Bayesian Dirichlet-Multinomial model. The left column shows posterior distributions for the open-weight cohort, and the right column shows posterior distributions for the closed-weight cohort. Rows correspond to the four severity tiers. Curves represent posterior kernel density estimates, dashed vertical lines indicate posterior means, and shaded regions represent 95% highest density intervals. This figure provides the complete comparator view supporting the focused main-text analysis.*

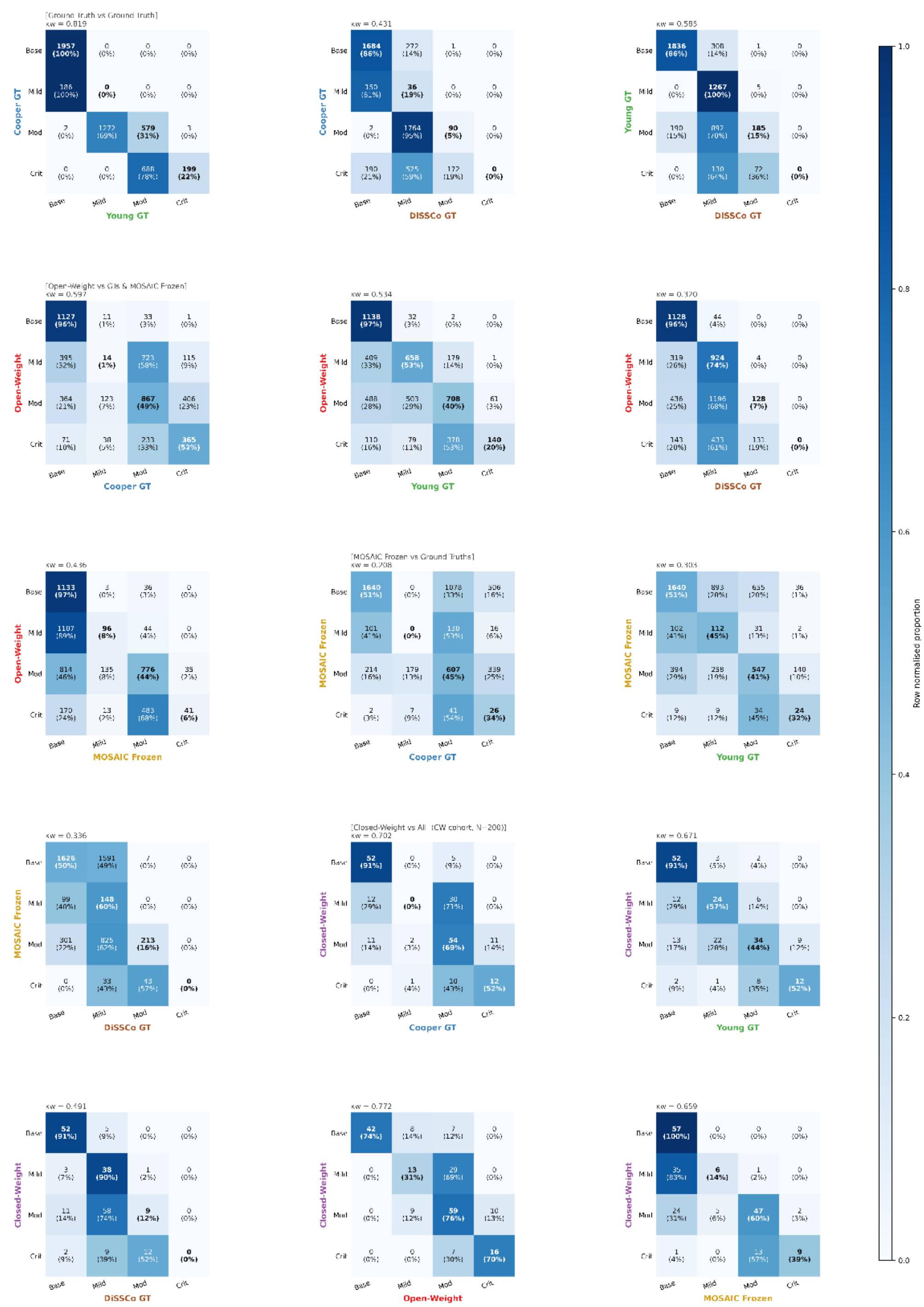

**Supplementary Figure 5.** Comprehensive row-normalised confusion matrices for all pairwise classifier comparisons.

***Legend:***Each panel shows a row-normalised confusion matrix comparing two classifiers across the four severity tiers (Baseline, Mild, Moderate, Critical). Rows represent the classifier named on the y-axis; columns represent the classifier named on the x-axis. Cell values report the absolute patient count and the row-normalised percentage, indicating the proportion of patients assigned to each column-classifier tier given a row-classifier assignment. Shading intensity reflects the row-normalised proportion, with darker cells indicating higher concentration. Quadratic weighted Cohen's κw is shown above each panel. Panels are grouped into four sections, colour-coded in the legend: Ground Truth vs Ground Truth (inter-GT reference comparisons, OW cohort N = 4,886); Open-Weight vs GTs and MOSAIC Frozen (OW cohort N = 4,886); MOSAIC Frozen vs Ground Truths (OW cohort N = 4,886); and Closed-Weight vs All (CW cohort N = 200). Diagonal dominance indicates strong agreement; off-diagonal mass concentrated at adjacent cells (Δ1 shifts) indicates bounded tier-boundary disagreement; off-diagonal mass at non-adjacent cells indicates more severe misclassification.

**Supplementary Table 9**. Severity tier assignment and posterior tier probabilities (CW cohort, N=200)

| Tier | Closed-Weight (base) | Young GT | Cooper GT | DiSSCo GT | MOSAIC Frozen | Open-Weight |
|---|---|---|---|---|---|---|
| **Baseline** | 57 (28.5%) | 79 (39.5%) | 75 (37.5%) | 68 (34.0%) | 117 (58.5%) | 42 (21.0%) |
| | π = 0.285 [0.220–0.344] | π = 0.392† | π = 0.372 | π = 0.339 | π = 0.579† | π = 0.211 |
| **Mild** | 42 (21.0%) | 50 (25.0%) | 3 (1.5%) | 110 (55.0%) | 11 (5.5%) | 30 (15.0%) |
| | π = 0.211 [0.159–0.269] | π = 0.250 | π = 0.020† | π = 0.544† | π = 0.059† | π = 0.152 |
| **Moderate** | 78 (39.0%) | 50 (25.0%) | 99 (49.5%) | 22 (11.0%) | 61 (30.5%) | 102 (51.0%) |
| | π = 0.387 [0.320–0.450] | π = 0.250† | π = 0.490† | π = 0.112† | π = 0.304 | π = 0.505† |
| **Critical** | 23 (11.5%) | 21 (10.5%) | 23 (11.5%) | 0 (0.0%) | 11 (5.5%) | 26 (13.0%) |
| | π = 0.118 [0.075–0.162] | π = 0.108 | π = 0.118 | π = 0.005† | π = 0.059† | π = 0.132 |

***Legend:*** *Each cell reports the number of patients assigned to that tier (n), the corresponding percentage of the cohort (%), and the posterior tier probability estimated from a Bayesian Dirichlet–Multinomial model (π). For CW MOSAIC, π is shown with its 95% highest density interval (HDI); for all comparator frameworks, π represents the posterior mean only. † indicates that the 95% HDI of the tier-probability difference (Δk = comparator − CW MOSAIC) excludes zero, confirming a systematic difference in how that tier is populated relative to CW MOSAIC.*

**Supplementary Table 10**. Baseline characteristics of the open-weight MOSAIC cohort stratified by severity tier at the $T_5$.

| Characteristic | Overall | Baseline | Mild | Moderate | Critical | Max SMD |
|---|---|---|---|---|---|---|
| **N** | 4,886 | 1,172 | 1,247 | 1,760 | 707 | — |
| **Age at T5, mean ± SD** | 45.6 ± 6.6 | 44.6 ± 6.5 | 45.2 ± 6.3 | 46.0 ± 6.7 | 46.7 ± 6.6 | 0.326 |
| **Sex** | — | — | — | — | — | — |
| **Female** | 2,516 (51.5%) | 593 (50.6%) | 664 (53.2%) | 901 (51.2%) | 358 (50.6%) | 0.053 |
| **Male** | 2,367 (48.4%) | 579 (49.4%) | 581 (46.6%) | 859 (48.8%) | 348 (49.2%) | 0.056 |
| **First drug class** | — | — | — | — | — | — |
| **Metformin** | 4,359 (89.2%) | 1,147 (97.9%) | 1,225 (98.2%) | 1,487 (84.5%) | 500 (70.7%) | 0.804 |
| **Insulin** | 524 (10.7%) | 25 (2.1%) | 20 (1.6%) | 272 (15.5%) | 207 (29.3%) | 0.804 |
| **Liraglutide (GLP-1)** | 3 (0.1%) | 0 (0.0%) | 2 (0.2%) | 1 (0.1%) | 0 (0.0%) | 0.057 |
| **DCSI score at T5, median [IQR]** | 1.0 [0.0–2.0] | 0.0 [0.0–0.0] | 1.0 [0.0–1.0] | 1.0 [0.0–2.0] | 2.0 [1.0–3.0] | 2.270 |
| **Complications at T5** | — | — | — | — | — | — |
| **Retinopathy** | 2,329 (47.7%) | 19 (1.6%) | 545 (43.7%) | 1,090 (61.9%) | 675 (95.5%) | 5.457 |
| **Nephropathy** | 1,363 (27.9%) | 4 (0.3%) | 39 (3.1%) | 709 (40.3%) | 611 (86.4%) | 3.503 |
| **Neuropathy** | 1,217 (24.9%) | 14 (1.2%) | 458 (36.7%) | 495 (28.1%) | 250 (35.4%) | 1.017 |
| **CVD** | 24 (0.5%) | 0 (0.0%) | 0 (0.0%) | 14 (0.8%) | 10 (1.4%) | 0.169 |
| **PVD** | 0 (0.0%) | 0 (0.0%) | 0 (0.0%) | 0 (0.0%) | 0 (0.0%) | 0.000 |
| **Cerebrovascular** | 106 (2.2%) | 0 (0.0%) | 4 (0.3%) | 62 (3.5%) | 40 (5.7%) | 0.346 |
| **Metabolic** | 0 (0.0%) | 0 (0.0%) | 0 (0.0%) | 0 (0.0%) | 0 (0.0%) | 0.000 |
| **Death in follow-up, n (%)** | 489 (10.0%) | 48 (4.1%) | 117 (9.4%) | 231 (13.1%) | 93 (13.2%) | 0.327 |
| **New complication in follow-up, n (%)** | 3,146 (64.4%) | 1,039 (88.7%) | 876 (70.2%) | 958 (54.4%) | 273 (38.6%) | 1.218 |

**Supplementary Table 11.** Person-years and crude incidence rates of the open-weight MOSAIC cohort.

| Severity | Outcome | N | FU Person-Years | N events | N censored | N competing deaths | IR per 100PY | 95% CI |
|---|---|---|---|---|---|---|---|---|
| **Overall** | New diabetes complication | 4886 | 18289.9 | 3146 | 1446 | 294 | 17.20 | [16.60–17.81] |
| **Overall** | All-cause mortality | 4886 | 22819.8 | 489 | 4397 | — | 2.14 | [1.96–2.34] |
| **Critical** | New diabetes complication | 707 | 2929.9 | 273 | 360 | 74 | 9.32 | [8.24–10.49] |
| **Critical** | All-cause mortality | 707 | 3239.7 | 93 | 614 | — | 2.87 | [2.32–3.52] |
| **Baseline** | New diabetes complication | 1172 | 4149.7 | 1039 | 118 | 15 | 25.04 | [23.54–26.61] |
| **Mild** | New diabetes complication | 1247 | 4616.9 | 876 | 302 | 69 | 18.97 | [17.74–20.27] |
| **Baseline** | All-cause mortality | 1172 | 5771.4 | 48 | 1124 | — | 0.83 | [0.61–1.10] |
| **Mild** | All-cause mortality | 1247 | 5936.2 | 117 | 1130 | — | 1.97 | [1.63–2.36] |
| **Moderate** | New diabetes complication | 1760 | 6593.4 | 958 | 666 | 136 | 14.53 | [13.62–15.48] |
| **Moderate** | All-cause mortality | 1760 | 7872.5 | 231 | 1529 | — | 2.93 | [2.57–3.34] |

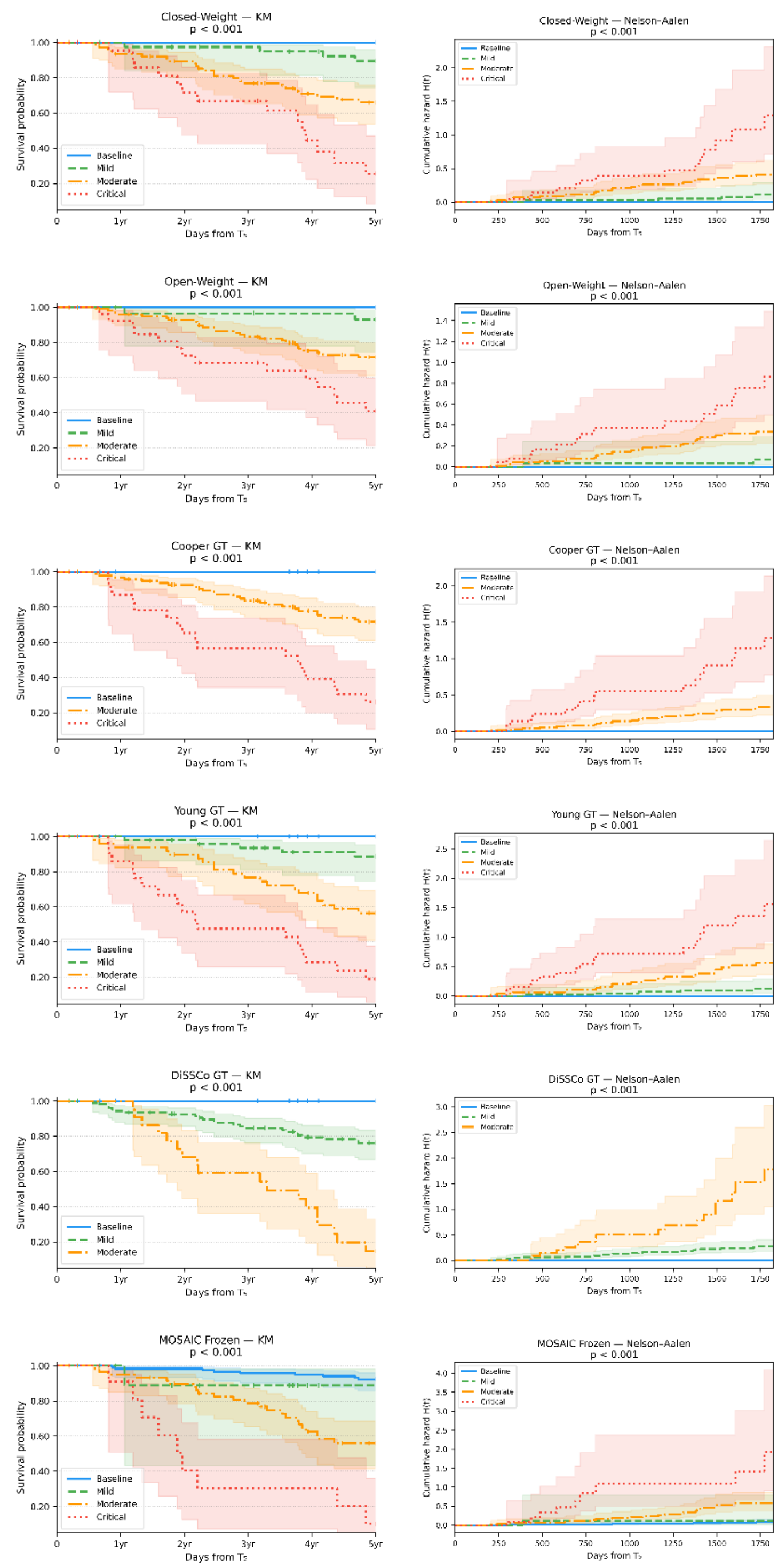

**Supplementary Figure 6.** Kaplan–Meier survival and Nelson–Aalen cumulative hazard curves for all-cause mortality for CW cohort.

*Kaplan–Meier survival curves and Nelson–Aalen cumulative hazard curves are shown for all-cause mortality during the five-year follow-up period after the* $T_5$ *landmark. Rows correspond to the classification framework used to assign severity tiers. The left column shows Kaplan–Meier survival probability over time, where lower curves indicate lower survival. The right column shows Nelson–Aalen cumulative hazard, where higher curves indicate greater accumulated mortality hazard. The x-axis represents days from the* $T_5$ *landmark, and the y-axis represents either survival probability or cumulative hazard, depending on the panel. Shaded areas indicate 95% confidence intervals. Log-rank tests are shown for each framework. Across all three systems, mortality risk differs significantly by severity tier.*

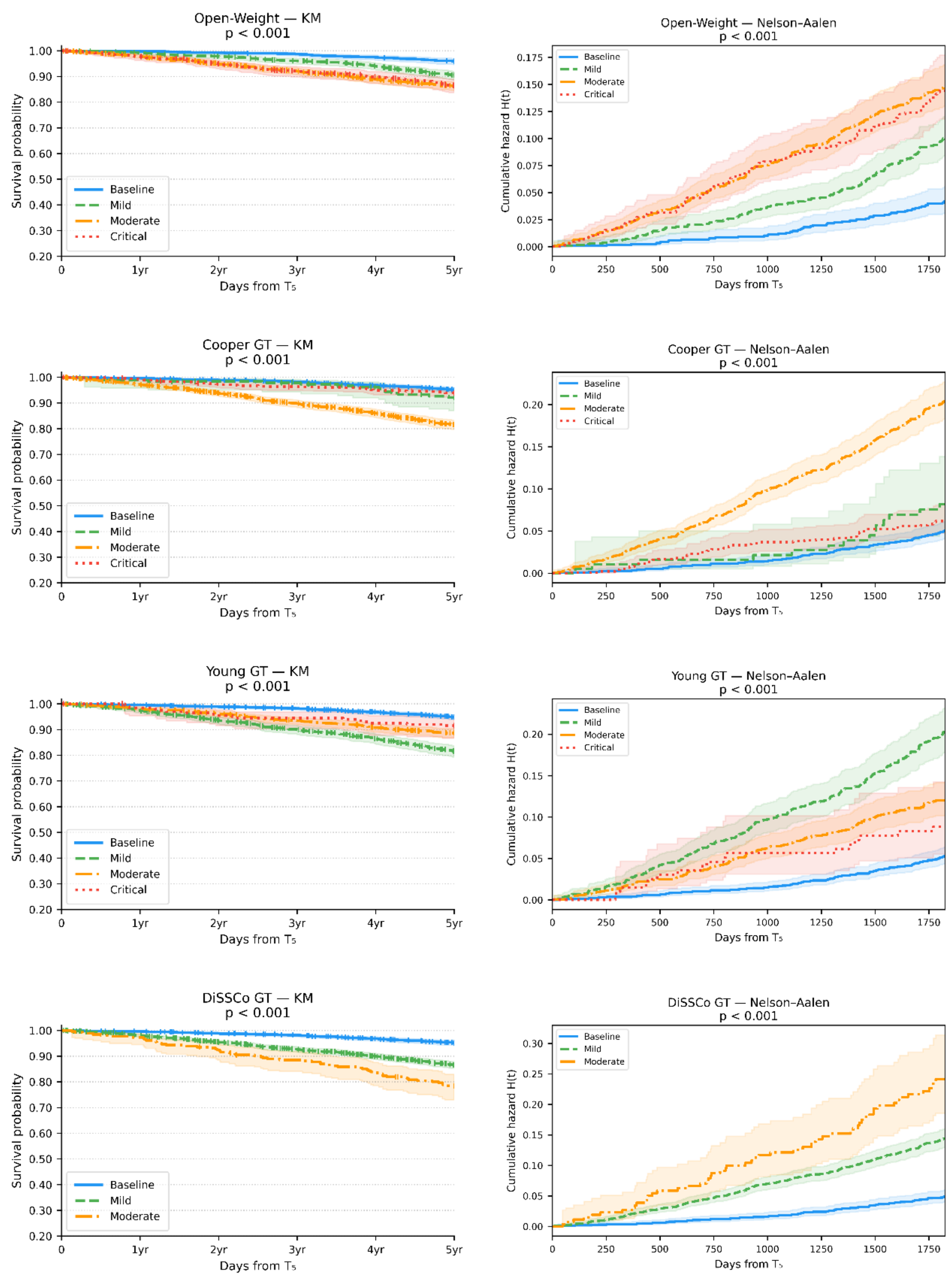

**Supplementary Figure 7.** Kaplan–Meier survival and Nelson–Aalen cumulative hazard curves for all-cause mortality for ground truths and mosaic frozen.

*Kaplan–Meier survival curves and elson–Aalen cumulative hazard curves are shown for all-cause mortality during the five-year follow-up period after the* $T_5$ *landmark. Rows correspond to the classification framework used to assign severity tiers. The left column shows Kaplan–Meier survival probability over time, where lower curves indicate lower survival. The right column shows Nelson–Aalen cumulative hazard, where higher curves indicate greater accumulated mortality hazard. The x-axis represents days from the* $T_5$ *landmark, and the y-axis represents either survival probability or cumulative hazard, depending on the panel. Shaded areas indicate 95% confidence intervals. Log-rank tests are shown for each framework. Across all three systems, mortality risk differs significantly by severity tier.*

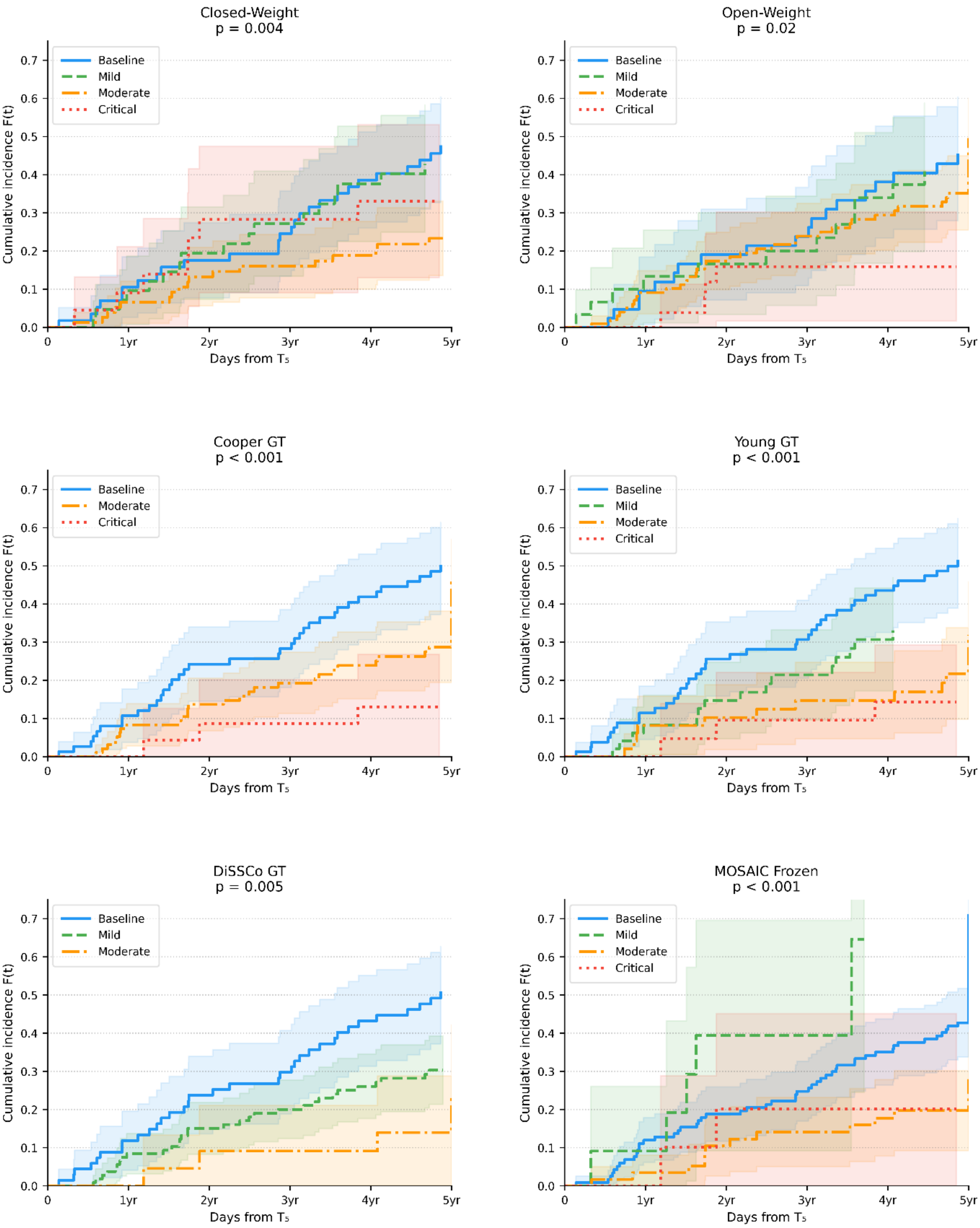

**Supplementary Figure 8.** Cumulative incidence of new diabetes complications by severity framework for CW cohort.

***Legend:*** *Cumulative incidence functions were estimated using the Aalen–Johansen estimator during the five-year follow-up period after the* $T_5$ *landmark, with death treated as a competing event.*

**New Diabetes Complication (Aalen-Johansen CIF) — Ground Truths (OW Cohort, N=4,886)**
**Cooper · Young · DiSSCo · MOSAIC Frozen · shared y-axis · zoomed**

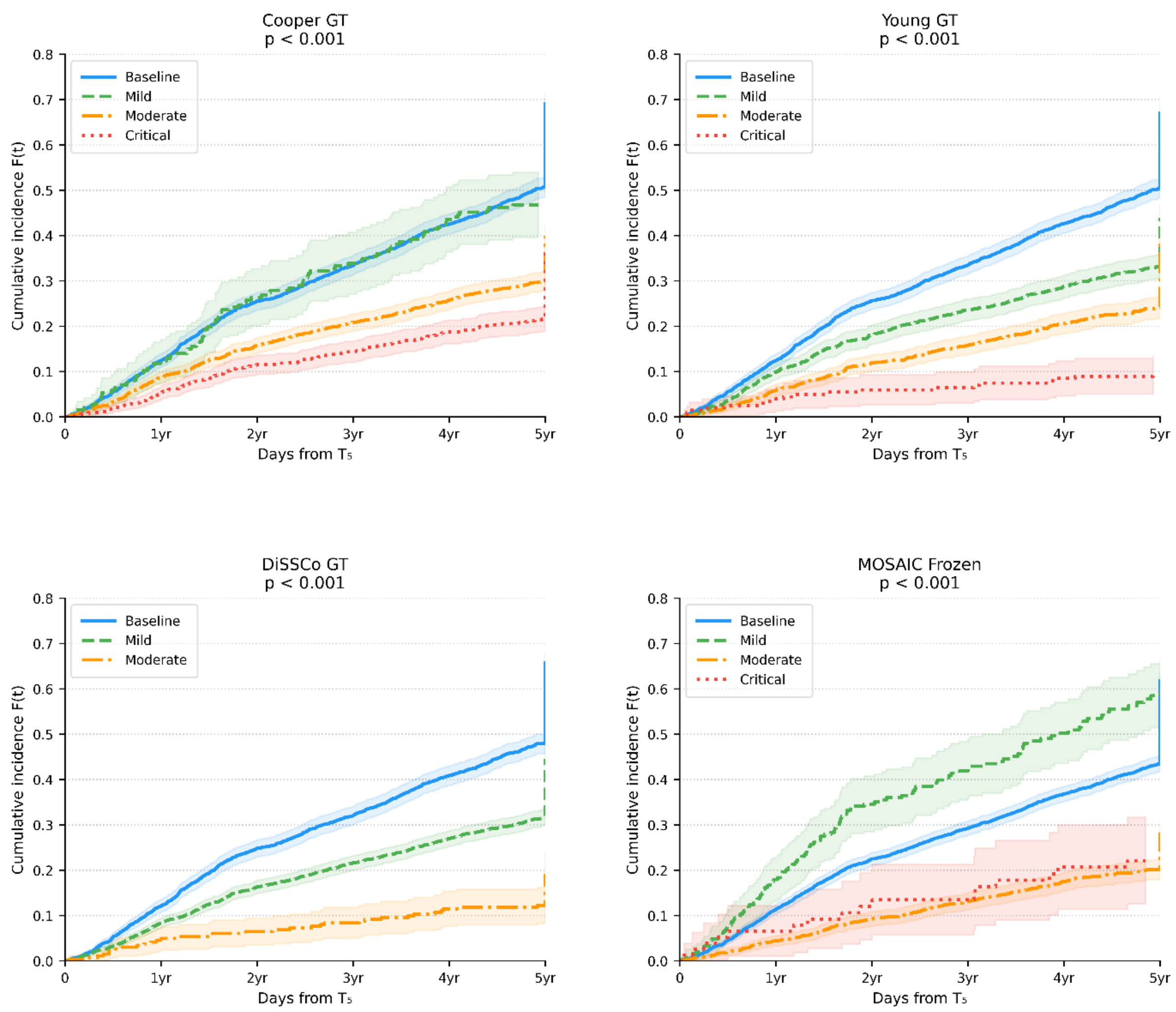

**Supplementary Figure 9.** Cumulative incidence of new diabetes complications by severity framework for OW cohort.

***Legend:*** *Cumulative incidence functions were estimated using the Aalen–Johansen estimator during the five-year follow-up period after the* $T_5$ *landmark, with death treated as a competing event.*

**Supplementary Table 12.** Illustrative discordance patterns: MOSAIC versus algorithmic ground-truth frameworks.

**PATTERN I**
**Insulin-Dependent Over-Classification via Pharmacotherapy Proxy**
*Patient a1733f62 — First-line insulin as sole severity trigger; zero complications or GT criteria met*

| PATIENT PROFILE | TIER ASSIGNMENTS | MOSAIC KEY EVIDENCE CITED | STRUCTURAL GAP ILLUSTRATED |
|---|---|---|---|
| **30.6 yr** Female<br>T2D diagnosis to $T_0$: 5 yr<br>No documented complications.<br>No lab data available.<br>No comorbidities. | CW MOSAIC **Moderate** ↑2<br>OW MOSAIC **Moderate** ↑2<br>Young GT **Baseline**<br>Cooper GT **Baseline** | Humulin 70/30 (premixed insulin) initiated as the first and only antidiabetic agent at $T_0$ and continued uninterrupted for the full 5-year window, meeting the Tier 3 Domain 7 pharmacotherapy proxy criterion for insulin dependence. No complications, comorbidities, or laboratory data were present to corroborate or contradict this signal. Clinically atypical presentation for T2D at age 30. | **Pharmacotherapy proxy without corroboration**<br>Neither Young GT nor Cooper GT incorporates treatment escalation as a severity criterion. Cooper GT requires coded complication evidence (comp_count = 0; insulin flag = True but insufficient). MOSAIC's Domain 7 pharmacotherapy proxy is triggered by insulin use alone — without requiring concurrent complication evidence — producing over-classification when insulin is prescribed for reasons other than established beta-cell failure (e.g., clinical preference or diagnostic uncertainty). |

**PATTERN II**
**Medication Reason-Code Inference of Cardiovascular Disease**
*Patient aa17ca13 — Simvastatin with CHD reason code inferred by MOSAIC; no formal ICD code on problem list*

| PATIENT PROFILE | TIER ASSIGNMENTS | MOSAIC KEY EVIDENCE CITED | STRUCTURAL GAP ILLUSTRATED |
|---|---|---|---|
| **55.9 yr** Male<br>T2D diagnosis to $T_0$: 5 yr<br>No coded complications.<br>Metformin monotherapy.<br>No formal CVD diagnosis code. | CW MOSAIC **Moderate** ↑2<br>OW MOSAIC **Moderate** ↑2<br>Young GT **Baseline**<br>Cooper GT **Baseline** | Simvastatin 20 mg prescribed with explicit reason code "Coronary Heart Disease" (SNOMED 53741008) at approximately 2.3 years post-$T_0$, confirming established macrovascular cardiovascular disease per Domain 5 Tier 3 criterion — despite the absence of a formal ICD-10 CHD code on the active problem list. Both CW and OW pipelines inferred CVD from this medication-linked reason code. Young GT and Cooper GT found no complications (comp_count = 0; retinopathy = None). | **Medication-linked surrogate diagnosis**<br>Algorithmic GT frameworks rely on condition-code tables (ICD-10 / SNOMED on the problem list) to detect CVD. When a diagnosis appears only as a medication reason code rather than a standalone condition entry, condition-code-only algorithms miss it entirely. MOSAIC reads across the medication record and infers the underlying condition from prescribing context. This may represent clinically valid evidence capture or over-inference from surrogate codes — a distinction that requires clinician adjudication. |

**PATTERN III**
**Two-Domain Concurrence Triggering Rule 0a Escalation**
*Patient 986dd9be — HbA1c ≥8.0% + insulin escalation within 11 months; mandatory Tier 4 not operationalised in any GT*

| PATIENT PROFILE | TIER ASSIGNMENTS | MOSAIC KEY EVIDENCE CITED | STRUCTURAL GAP ILLUSTRATED |
|---|---|---|---|
| **44.0 yr** Female<br>T2D diagnosis to $T_0$: 5 yr<br>Retinopathy (NPDR).<br>Diabetic renal disease.<br>HbA1c 10.3% at $T_0$, 8.4% at 12 months.<br>Stage 2 hypertension (vitals). | CW MOSAIC **Critical** ↑1<br>OW MOSAIC **Moderate**<br>Young GT **Moderate**<br>Cooper GT **Moderate**<br>*CW assessors disagreed (1× Moderate, 1× Critical); consolidated to Critical.* | Two independent domain criteria simultaneously reached Tier 3: Domain 1 — HbA1c 10.3% at $T_0$ and 8.4% at 12 months (both ≥8.0% Tier 3 threshold), and Domain 7 — escalation from metformin + liraglutide to insulin within 11 months of $T_0$ (confirming beta-cell failure). Per Framework Rule 0a, concurrent Tier 3 signals across two independent domains trigger mandatory escalation to Tier 4 (Critical). Corroborated by coded retinopathy, renal disease, hypertension, and severe dyslipidemia. OW MOSAIC classified Moderate (Tier 3), not triggering Rule 0a. | **Multi-domain escalation rule (Rule 0a)**<br>Young GT and Cooper GT assigned Moderate based on retinopathy and renal disease — clinically appropriate and consistent with OW MOSAIC. MOSAIC's Rule 0a (≥2 concurrent Tier 3 domain triggers → automatic Tier 4) has no equivalent in any GT framework. This rule encodes a specific clinical heuristic — that multi-system deterioration signals greater severity than any single domain — that the algorithmic GTs do not operationalise. Whether this represents genuine over-classification or a clinically valid escalation that GTs lack the granularity to capture cannot be determined from algorithmic comparisons alone. |